\documentclass[10pt]{article} % For LaTeX2e
\usepackage[accepted]{tmlr}

\usepackage{microtype}
\usepackage{graphicx}
\usepackage{subcaption}
\usepackage{booktabs} % for professional tables
\usepackage[hidelinks]{hyperref}

\usepackage{amsmath}
\usepackage{amssymb}
\usepackage{mathtools}
\usepackage{amsthm}

\usepackage[capitalize,noabbrev]{cleveref}

\usepackage{multirow}
\usepackage{xcolor}
\usepackage{amsmath,amsfonts,bm, bbm}

\newcommand{\mH}{\mathcal{H}}
\newcommand{\mX}{\mathcal{X}}
\newcommand{\mY}{\mathcal{Y}}

\newtheorem{theorem}{Theorem}

\newtheorem{lemma}[theorem]{Lemma}
\newtheorem{proposition}[theorem]{Proposition}
\newtheorem{assumption}{Assumption}
\theoremstyle{definition}
\newtheorem{definition}{Definition}
\newtheorem{remark}{Remark}

\allowdisplaybreaks

\def\1{\bm{1}}

\DeclareMathAlphabet{\mathsfit}{\encodingdefault}{\sfdefault}{m}{sl}
\SetMathAlphabet{\mathsfit}{bold}{\encodingdefault}{\sfdefault}{bx}{n}

\def\gH{{\mathcal{H}}}

\def\gS{{\mathcal{S}}}

\def\gZ{{\mathcal{Z}}}

\newcommand{\E}{\mathbb{E}}

\newcommand{\R}{\mathbb{R}}

\newcommand{\dd}{\,\mathrm{d}}
\newcommand{\ip}[1]{\left\langle #1 \right\rangle}
\newcommand{\nor}[1]{\left\lVert #1 \right\rVert}

\newcommand{\blue}[1]{#1}

\usepackage[textsize=tiny]{todonotes}

\title{Theoretical Refinement of CLIP by Utilizing Linear Structure of Optimal Similarity}

\author{\name Naoki Yoshida\thanks{Equal contribution.}\hspace{0.4em}\thanks{Work done during an internship at Sony Group Corporation.}
\email n-yoshida@g.ecc.u-tokyo.ac.jp \\
\addr The University of Tokyo
\AND
\name Satoshi Hayakawa\footnotemark[1]\hspace{0.4em}\thanks{Currently at The University of Tokyo. Work done while at Sony Group Corporation.}
\email hayakawa@mist.i.u-tokyo.ac.jp \\
\addr Sony Group Corporation
\AND
\name Yuhta Takida
\email yuta.takida@sony.com \\
\addr Sony Group Corporation
\AND
\name Toshimitsu Uesaka
\email toshimitsu.uesaka@sony.com \\
\addr Sony Group Corporation
\AND
\name Hiromi Wakaki
\email hiromi.wakaki@sony.com \\
\addr Sony Group Corporation
\AND
\name Yuki Mitsufuji
\email yuhki.mitsufuji@sony.com \\
\addr Sony Group Corporation
}

\def\month{09}  % Insert correct month for camera-ready version
\def\year{2026} % Insert correct year for camera-ready version
\def\openreview{\url{https://openreview.net/forum?id=xRKVzuZ68J}} % Insert correct link to OpenReview for camera-ready version

\begin{document}

\maketitle

\begin{abstract}
In this study, we propose an enhancement to the similarity computation mechanism in multimodal contrastive pretraining frameworks such as CLIP. 
Prior theoretical research has demonstrated that the optimal similarity metrics between paired modalities should correspond to the pointwise mutual information (PMI) between the two modalities.
However, the current implementations of CLIP and its variants fail to fully utilize the underlying linear structure of PMI.
We therefore propose KME-CLIP, which leverages this structure through the inner product in a reproducing kernel Hilbert space \blue{(RKHS). We theoretically prove that, under our assumptions, the KME-CLIP similarity can bring the contrastive loss arbitrarily close to its optimal value, which is attained by PMI, as the size of the point set grows, and we empirically evaluate KME-CLIP against CLIP and its kernel-based variants across several retrieval and classification tasks.}
\end{abstract}

\section{Introduction}

{
A central concern in multimodal contrastive learning frameworks such as CLIP \citep{radford2021learning} is to improve their accuracy through obtaining representations that precisely align with the underlying structure of data. 
For this purpose, several studies have continued to improve CLIP through modification of loss functions \citep{goel2022cyclip,mu2022slip} and similarity functions \citep{desai2023hyperbolic,uesaka2025wpse}.

For improving the accuracy of multimodal contrastive learning, theoretical results suggest that the similarity function should approximate the pointwise mutual information (PMI), which is defined by $\log(p(x,y)/(p(x)p(y)))$ for two datapoints $x$ and $y$ from different modalities.
In fact, the similarity function minimizing the population contrastive loss is known to be PMI \citep{oord2018representation, zhang2023deep}, and closer approximation of PMI has been shown to yield improvements across downstream tasks including retrieval and zero-shot classification \citep{oko2025multimodal}.

Despite such an importance of PMI, the similarity function of CLIP and its variants have overlooked a fundamental structural property of PMI: under a standard conditional independence assumption, the exponential of PMI possesses a linear structure as an inner product in $L^2$ space. Neither CLIP nor its variants exploit this linear structure, and ignoring it leads to failure even in a simple toy example.

In this paper, we propose KME-CLIP (Kernel Mean Embedding CLIP), a similarity function that improves PMI approximation by explicitly exploiting this linear structure. 
KME-CLIP projects embeddings into a reproducing kernel Hilbert space (RKHS), where the linear structure of inner products in $L^2$ space can be naturally captured through its Hilbert-space inner products.
Specifically, embeddings from image and text encoders are represented as positive functions in the RKHS, and their similarity is computed as the logarithm of their RKHS inner product, enabling more principled and effective PMI approximation.
}

Our contributions are summarized as follows:
\begin{enumerate}
    \item We identify that PMI exhibits an inherent linear structure under reasonable assumptions regarding data distributions
    (\Cref{section: intuition}),
    which remains unexploited in conventional CLIP.
    We then propose KME-CLIP, whose similarity leverages this linearity through RKHS and harnesses the functional representation capacity of reproducing kernels (\Cref{section: kme-clip}).
    % which replaces the similarity of CLIP and has a high expressive power,
    % utilizing a rich functional representation of RKHS.
    \item We theoretically establish that the similarity metric in KME-CLIP can approximate PMI with arbitrary accuracy
    when the  size of point set is sufficiently increased (\Cref{section: theory_of_our_method}). Furthermore, we demonstrate theoretically that CLIP faces \blue{a structural limitation} in approximating PMI \blue{in a low-dimensional setting} (\Cref{section: theory_clip}).
    \item In our experiments (\Cref{section: experiment}), we train CLIP, WPSE~\citep{uesaka2025wpse}, and
    KME-CLIP on CC3M \citep{sharma-etal-2018-conceptual} and CC12M
    \citep{changpinyo2021conceptual} under the same experimental setup.
    KME-CLIP improves over the baselines on average in image--text retrieval
    benchmarks using CC3M, MSCOCO \citep{cococaption}, and Flickr30K
    \citep{flikr}, and achieves the best average zero-shot classification
    performance across 13 datasets, including ImageNet
    \citep{russakovsky2015imagenet} and CIFAR-100
    \citep{krizhevsky2009learning}.
    We also report linear-classification results,
    point-set-size ablations, computational-cost comparisons, and modality-gap
    diagnostics.
    \blue{Beyond these downstream evaluations, we directly measure PMI fidelity
    in a synthetic setting where PMI is known analytically, using a matched
    control (KME-no-log) that isolates the outer logarithm: the logarithm is
    the more faithful and better-ranking choice there, whereas the control is
    stronger on CC3M downstream metrics
    (Section~\ref{sec: outer-log}).}
\end{enumerate}

\section{Preliminaries}
\subsection{Contrastive language-image pre-training (CLIP)}
CLIP \citep{radford2021learning} trains image encoder and text encoder to maximize the cosine similarity between embeddings of the corresponding image and caption pairs.
This method aims to learn the representation which projects the paired image and caption to aligned embedding vectors.

While we refer to them as image and text for simplicity, our arguments hold for any pair of modalities.
Let $g^{\mX}:\mX\rightarrow\R^d$ and $g^{\mY}:\mY\rightarrow\R^d$ be
$L^2$-normalized image and text encoders, respectively.
Given a minibatch of image dataset and the corresponding caption dataset $\{(x_i, y_i)\}_{i=1}^n$,
% Let the similarity function between the image embedding $u$ and the text embedding $v$ be $S(u,v)$.
the training loss, which we call the CLIP loss, is calculated as follows:
% \begin{align}\label{loss_of_clip}
%     \frac12\bigg[
%     &-\frac{1}{2n}\sum_{i=1}^n \log\left(\frac{\exp(g^{\mX}(x_i)^\top g^{\mY}(y_i)/\tau)}{\sum_{j=1}^n \exp(g^{\mX}(x_i)^\top g^{\mY}(y_j)/\tau)}\right)
%     \\&
%     -\frac{1}{2n}\sum_{i=1}^n \log\left(\frac{\exp(g^{\mX}(x_i)^\top g^{\mY}(y_i)/\tau)}{\sum_{j=1}^n \exp(g^{\mX}(x_j)^\top g^{\mY}(y_i)/\tau)}\right)\bigg].
% \end{align}
\begin{align*}%\label{loss_of_clip}
    \hspace{-0.5pt}
    \frac12\Biggl[
    -\frac{1}{n}\sum_{i=1}^n \log\Biggl(\frac{\exp(\frac1\tau g^{\mX}(x_i)^\top g^{\mY}(y_i))}{\sum_{j=1}^n \exp(\frac1\tau g^{\mX}(x_j)^\top g^{\mY}(y_i))}\Biggr)
    -\frac{1}{n}\sum_{i=1}^n \log\Biggl(\frac{\exp(\frac1\tau g^{\mX}(x_i)^\top g^{\mY}(y_i))}{\sum_{j=1}^n \exp(\frac1\tau g^{\mX}(x_i)^\top g^{\mY}(y_j))}\Biggr)\Biggr].
\end{align*}
Here, $\tau$ denotes the temperature parameter, which we consider to be a learnable parameter in this paper, controlling the influence of inner product magnitudes on the loss function.
% Moreover, the expected value of the loss function over the distribution of the dataset is written as
% \begin{align}\label{exp_of_clip}
%     &\frac{1}{2} \E_{(x,y)}\bigg[
%         - \log \frac{\exp(f^{\mX}(x)^\top f^{\mY}(y)/\tau)
%         }{
%         \E_{x'}[\exp(f^{\mX}(x')^\top f^{\mY}(y)/\tau)]
%         }\bigg] \\
%     +   &\frac{1}{2} \E_{(x,y)}\bigg[
%         - \log \frac{\exp(f^{\mX}(x)^\top f^{\mY}(y)/\tau)
%         }{
%         \E_{y'}[\exp(f^{\mX}(x)^\top f^{\mY}(y')/\tau)]
%         }\bigg]
% \end{align}

The CLIP loss is a proxy of the performance on the image-to-text and text-to-image retrieval tasks:
for instance, the first term is measuring the probability that we retrieve $y_i$ among $\{y_j\}_{j=1}^n$
based on the distribution induced from the logits $\{g^{\mX}(x_i)^\top g^{\mY}(y_j)/\tau\}_{j=1}^n$.
Thus, the minimization of the training loss implies high performance in cross-modal retrieval tasks,
which is one of the primary tasks where CLIP embeddings are used. Indeed, theoretical analysis demonstrates that when the CLIP loss approaches its minimum, CLIP's performance on zero-shot classification tasks improves significantly \citep{oko2025multimodal}.
%{\color{red}[Needs modification: what are retrieval tasks?]}

% The minimization of the training loss directs to the high performance in a retrieval task, which is one of the downstream tasks in CLIP.
% Image-to-text retrieval is a task that, given a collection of ground truth image-caption pairs, aims to retrieve the caption with the highest similarity to each query image based on their embeddings, and evaluates whether the retrieved caption correctly matches to the corresponding caption. Text-to-image retrieval is the reverse of it, where the roles of images and captions are interchanged.
% Minimizing \eqref{loss_of_clip} results in increasing the similarity of paired image and text and decreasing the similarity of not paired ones. As a result, the probability of retrieving the correct match increases while the probability of retrieving incorrect matches decreases, thereby improving retrieval accuracy on the training data.

\subsection{Pointwise mutual information as best similarity metric}\label{section: PMI}
In this section, we consider a population loss variant of the CLIP loss.
Let $X$ and $Y$ be $\mX$- and $\mY$-valued random variables, representing image and text, respectively. Throughout our analysis, we assume the existence of density functions for both $X$ and $Y$.
\begin{assumption}[Density function]\label{assumption: density}
    We assume that $X$ and $Y$ have probability marginal density functions $p(x)$ and $p(y)$
    as well as a joint density $p(x,y)$ with respect to reference measures
    $\nu_X$, $\nu_Y$, and $\nu_{X}\otimes\nu_{Y}$, respectively.
\end{assumption}

If we generalize from the scaled similarity of CLIP $g^{\mX}(x)^\top g^{\mY}(y)/\tau$ to any similarity metric $S(x,y)$, the training loss (minus $\log n$) converges to the following as we increase the minibatch size $n$:
\begin{align}\label{exp_of_loss}
    L_S := \frac{1}{2} \mathbb{E}_{p(x,y)}\bigg[
    - \log \frac{\exp(S(x,y))
    }{
    \mathbb{E}_{p(x')}[\exp(S(x',y))]
    }\bigg]+\frac{1}{2} \mathbb{E}_{p(x,y)}\bigg[
    - \log \frac{\exp(S(x,y))
    }{
    \mathbb{E}_{p(y')}[\exp(S(x,y'))]
    }\bigg].
\end{align}
We note that minimizing $L_S$ directly enhances CLIP's performance in image-to-text and text-to-image retrieval tasks, analogous to the effect of minimizing the standard CLIP loss.

Remarkably, the optimal solution $S(x,y)$ of $L_S$ is known theoretically.
We first define pointwise mutual information as the probabilistic relationship between two modalities.
\begin{definition}[Pointwise mutual information]\label{definition: pmi}
    %Let $(X, Y)\sim p(x, y)$ be a paired random variable.
    For $X, Y$ under \Cref{assumption: density},
    the \textit{pointwise mutual information} (PMI) between them
    at points $x\in\mX$ and $y\in\mY$ is defined as 
    %\begin{align}
    $\text{PMI}(x,y):=\log\frac{p(x,y)}{p(x)p(y)}$.
    %    \label{eq: pmi-def}
    %\end{align}
\end{definition}
% {\color{red}[add context??]}
We now establish that the optimal similarity function $S(x,y)$ that minimizes $L_S$ corresponds to PMI, as formalized in the following theorem.
\begin{theorem}[Proposition1, \cite{zhang2023deep}]\label{thm: pmi-minimzer}
$L_S$ in \eqref{exp_of_loss} attains its minimum
if we have $S(x,y)=\text{\rm PMI}(x, y) +\mathrm{const}$ for all $x\in\mX$ and $y\in\mY$.
\end{theorem}
It is also known that the minimum of $L_S$ is \blue{$-I(X;Y)$, where $I(X;Y)$ is the mutual information between $X$ and $Y$.}
From \Cref{thm: pmi-minimzer}, constructing a similarity function that aligns with PMI is an important direction for improving the accuracy of contrastive learning. Motivated by this, we investigate what kinds of similarity functions can satisfy this requirement in the next section.

{
\subsection{RKHS notation and finite kernel embeddings}

We introduce the RKHS notation needed to define KME-CLIP.
Let $k:\R^d\times\R^d\to\R$ be a positive definite kernel and let
$\mH$ be its associated reproducing kernel Hilbert space.
We write
$
    \phi(u):=k(u,\cdot)\in\mH
$
for the canonical feature map. The reproducing property gives
\[
    \ip{\phi(u),\phi(v)}_\mH
    =
    \ip{k(u,\cdot),k(v,\cdot)}_\mH
    =
    k(u,v).
\]
Thus, inner products in $\mH$ linearize non-linear relationships between $u$ and $v$,
and they can be computed through kernel evaluations.
Both are desired properties for efficiently implementing the linear structure discussed in Section~\ref{section: intuition}.

KME-CLIP represents each sample by a finite nonnegative measure over token embeddings.
For a finite nonnegative measure
$
    \mu=\sum_{i=1}^m w_i\delta_{u_i}
$ with $
    w_i\ge 0$ and $u_i\in\R^d$,
its kernel (mean) embedding is
\[
    \Phi(\mu)
    :=
    \int_{\R^d} k(u,\cdot)\dd\mu(u)
    =
    \sum_{i=1}^m w_i k(u_i,\cdot)
    \in\mH.
\]
For two such measures
$\mu=\sum_{i=1}^{m_1} w_i\delta_{u_i}$ and
$\nu=\sum_{j=1}^{m_2} w'_j\delta_{v_j}$, we have
\[
    \ip{\Phi(\mu),\Phi(\nu)}_\mH
    =
    \sum_{i=1}^{m_1}\sum_{j=1}^{m_2}
    w_iw'_j k(u_i,v_j).
\]
This finite weighted form is exactly the form used in Section~\ref{section: kme-clip}: the image and text encoders output weighted point sets, and their similarity is defined as the logarithm of the RKHS inner product between the corresponding finite kernel embeddings.
}

\section{Proposed method}

\subsection{Key insight: linear structure in $\exp(\text{PMI})$}\label{section: intuition}
Our key insight is that $\exp(\text{PMI})$ can be expressed as an inner product in $L^2$ space under certain conditions,
where $\exp(\text{PMI})$ simply refers to the exponential of PMI from \Cref{definition: pmi}.
Before formalizing this insight, we introduce the following assumption for technical reasons.
\begin{assumption}[Conditional independence]\label{assumption: conditional}
    We assume that there exists a latent variable $Z$ taking values in a compact metric space $\gZ$, with a density function $\rho(z)$ with respect to a reference (Borel) measure $\nu$.
    We further assume that the conditional densities satisfy
    $
        p(x,y|z)=p(x|z)p(y|z).
    $
    % We also assume that $\gZ$ is compact and $\gZ = \mathop\mathrm{supp}\rho$ holds; otherwise we can re-take the support as the latent space.
\end{assumption}

{
% \begin{remark}
%     Assumption \ref{assumption: density} and \ref{assumption: conditional} is about conditional independence, which is often adopted in theoretical studies of multimodal representation learning \citep{chen2024understanding, oko2025multimodal}. Intuitively, this assumes the existence of a latent variable that represents the underlying topic of images and texts.
%     This assumption is not an unrealistic assumption, but rather a realistic description of actual data-generation processes commonly used in practice.
%     Indeed, most text-image datasets are generated in one of two ways: (a) adding prompts to given images (e.g., MSCOCO), or (b) collecting images corresponding to each class label (e.g., CIFAR-10, ImageNet). In case (a), the data-generation process can naturally be viewed as conditionally independent. Specifically, we sample image data $x$ from $p(x)$ and text data $y$ according to $p(y|x)$ in this case. Hence, when we set $z=x$, we have 
%     \[
%         p(x,y|x)=p(y|x)=p(x|x)p(y|x),
%     \]
%     which represents the conditional independence by a latent $x$. Case (b) represents the opposite case.
% \end{remark}
}
{
\begin{remark}
    \blue{Assumptions \ref{assumption: density} and \ref{assumption: conditional} together posit a latent variable $Z$ under which the two modalities are conditionally independent. This is a standard modeling idealization in theoretical studies of multimodal representation learning \citep{chen2024understanding, oko2025multimodal}, where $Z$ is interpreted as the underlying topic shared by an image and its caption.
    Conditional independence alone is not restrictive: taking $\gZ=\mX$ and $z=x$ satisfies $p(x,y| z)=p(x|z)p(y| z)$ for any joint distribution, so it carries no content by itself. The substantive content of our analysis comes from combining it with the regularity assumption (\Cref{assumption: lipschitz}): the degenerate choice $z=x$ would generally violate the uniform Lipschitz condition on $p(x| z)/p(x)$. Our assumptions should therefore be read as requiring a \emph{low-complexity} shared latent over which the conditional density ratios vary smoothly.}
\end{remark}

}

Under Assumption \ref{assumption: density} and \ref{assumption: conditional}, $\exp(\text{PMI})$ is equivalent to an inner product in an $L^2$ space:
\begin{align*}
    \frac{p(x, y)}{p(x)p(y)}
    &=\frac1{p(x)p(y)}\int p(x|z)p(y|z)\dd\rho(z)
    = \ip{\frac{p(x|\cdot)}{p(x)}, \frac{p(y|\cdot)}{p(y)}}_{L^2(\rho)},
\end{align*}
where $\mathrm{d}\rho(z)=\rho(z)\mathrm{d}\nu(z)$ gives a probability measure over $\gZ$.

Thus, a straightforward way to interpret $\exp(\text{PMI})$ is that
it is an inner product between two embedding functions in the $L^2$ space.
However, CLIP tries to approximate this ``linear'' object by
$\exp(g^{\mX}(x)^\top g^{\mY}(y)/\tau)$, which is inherently nonlinear due to the exponential operation.
% a natural way to express $\exp(\text{PMI})$ by the similarity between two vectors is
% natural way to express $\exp(\text{PMI})$ by the similarity between two vectors is
% using an inner product between vectors in some linear space.
% the infinite dimensional vectors and calculating the similarity by the inner product of them.
% the logarithm of the inner product of them.

To exploit the linearity structure of $\exp(\text{PMI})$,
we consider projecting the embedding vectors onto an RKHS, which is an infinite dimensional linear space, and defining the logarithm of the inner product in the RKHS as the similarity metric corresponding to $S(x, y)$.

\subsection{Proposed method: KME-CLIP}\label{section: kme-clip}
From the discussion in the previous section, we propose \textit{kernel mean embedding CLIP} (KME-CLIP), which approximates the $L^2$ inner product by an inner product in the RKHS (see also Section \ref{section: implementation} for the implementation).
Let $k$ be a symmetric positive definite kernel on $\R^d$ and $\mH$ be the associated RKHS.
We further assume the positivity of the kernel ($k>0$) for implementation reasons.
The image embedding function of KME-CLIP has the following components ($\R_+$ is the set of positive real numbers):
\begin{itemize}
    \item Encoders: $f_i^\mX:\mX\to\R^d$ for $i=1,\ldots, m_\mX$.
    \item Positive weight functions: $w_i^\mX:\mX\to\R_+$ for $i=1,\ldots, m_\mX$.
\end{itemize}
We refer to the multiple embeddings generated from the several encoders $f_i^\mX$ as a \emph{point set} and define the number of encoders, $m_\mX$, as its size.
These encoders and weight functions are learnable, while the size of point set $m_\mX$ is fixed as a hyper-parameter.
Similarly, KME-CLIP has their text-side counterpart:
$m_\mY$, $f_i^\mY$, and $w_i^\mY$.
Given the kernel $k$, RKHS $\mH$, and these components,
we embed $x\in\mX$
and $y\in\mY$
into $\mH$ as follows:
\begin{align}
    &h_\theta^\mX(x):=\sum_{i=1}^{m_\mX}w_i^\mX(x)k(f_i^\mX(x), \cdot)\in\mH,
    \qquad h_\theta^\mY(y)
    :=\sum_{j=1}^{m_\mY}w_j^\mY(y)k(f_j^\mY(y),\cdot)\in\mH.
    \label{eq: kme-embedding}
\end{align}
% Technically, with the necessity to keep the content within the logarithm positive, KME-CLIP projects the embedding vectors from several encoders in each modal $\{f^\mX_i:\mX\rightarrow\R^d\}_i$ and $\{f^\mY_j:\mY\rightarrow\R^d\}_j$ onto RKHS with a positive kernel function $k:\R^d\times\R^d\rightarrow\R_+$ and positive weight functions $\{w_i^{\mX}:\mX\rightarrow\R_+\}_i, \{w_j^{\mY}:\mY\rightarrow\R_+\}_j$, where $\R_+$ denotes the set of positive real numbers. 
% % Furthermore, since $\exp(\text{PMI})=\frac{p(x,y)}{p(x)p(y)}$ may become $0$, the kernel function $k$ should tend to take values close to zero, such as the Gaussian kernel. 
% Then, the logarithm of their inner product in RKHS represents the similarity (see also Figure \ref{fig: method} for summarizing our method):
% \[
%     g^{\mX}(x)=\sum_i w_i^{\mX}(x) k(f_i^{\mX}(x),\cdot),
%     \qquad
%     g^{\mY}(y)=\sum_j w_j^{\mY}(y) k(\cdot, f_j^{\mY}(y)),
% \]
Then we define the logarithm of their inner product in $\mH$ as our proposed similarity metric:
\begin{align}
    S(x,y)=\log\ip{h_\theta^\mX(x), h_\theta^\mY(y)}_\mH
    %&\log\ip{\sum_{i=1}^{m_\mX} w_i^{\mX}(x) k(f_i^{\mX}(x),\cdot), \sum_{j=1}^{m_\mY} w_j^{\mY}(y) k(\cdot, f_j^{\mY}(y))}_{\mH} \notag\\
    &=\log
    %\sum_{\substack{1\le i\le m_\mX \\1\le j\le m_\mY}}
    \sum_{i=1}^{m_\mX}
    \sum_{j=1}^{m_\mY}
    w_i^{\mX}(x)w_j^{\mY}(y) k(f_i^{\mX}(x), f_j^{\mY}(y)),
    \label{similarity_kernel}
\end{align}
where $\ip{\cdot,\cdot}_\mH$ denotes the inner product in $\mH$.
Since the inside of the logarithm must be positive,
we have constrained $k$ and weight functions to be positive.
% Because of the positivity of weight functions, 
% $\mu_x:=\sum_{i=1}^{m_\mX}w_i^\mX(x)\delta_{f_i^\mX(x)}$,
% where $\delta_a$ is the delta distribution at point $a\in\R^d$,
% becomes a positive measure.
% Then, the embedding in \eqref{eq: kme-embedding} is rewritten as
% $x\mapsto \int k(z, \cdot)\dd\mu_x(z)$,
% which is called kernel mean embedding when $\mu_x$ is a probability measure \citep{muandet2017kernel}.
% This is why our method is named KME-CLIP. 

By choosing an appropriate kernel $k$,
we can prove that the similarity metric with the form \eqref{similarity_kernel} is capable of approximating PMI with arbitrary accuracy as we increase $m_\mX$ and $m_\mY$ (see Section \ref{section: theory}).

\subsection{Implementation}\label{section: implementation}
To generate multiple point set image embeddings $\{f_i^\mX(x)\}_{i=1}^{m_{\mX}}$, we leverage the intermediate features from Vision Transformer \citep{dosovitskiy2020image}, which naturally produces multiple token outputs.
To derive the positive weights $\{w_i^\mX(x)\}_{i=1}^{m_{\mX}}$, we similarly utilize these intermediate features by mapping them to additional features and applying an activation function to ensure positivity. In the same way as CLIP, we apply $L^2$-normalization for each image embedding $f_i^\mX(x)$.

For the text embeddings, we utilize multiple output tokens from the Transformer architecture \citep{vaswani2017attention}.
Regarding the text positive weights $\{w_i^\mY(y)\}_{i=1}^{m_{\mY}}$, we process the Transformer outputs using the same methodology applied to image embeddings.

\begin{figure*}[htbp]
\begin{center}
\includegraphics[width=1.\linewidth]{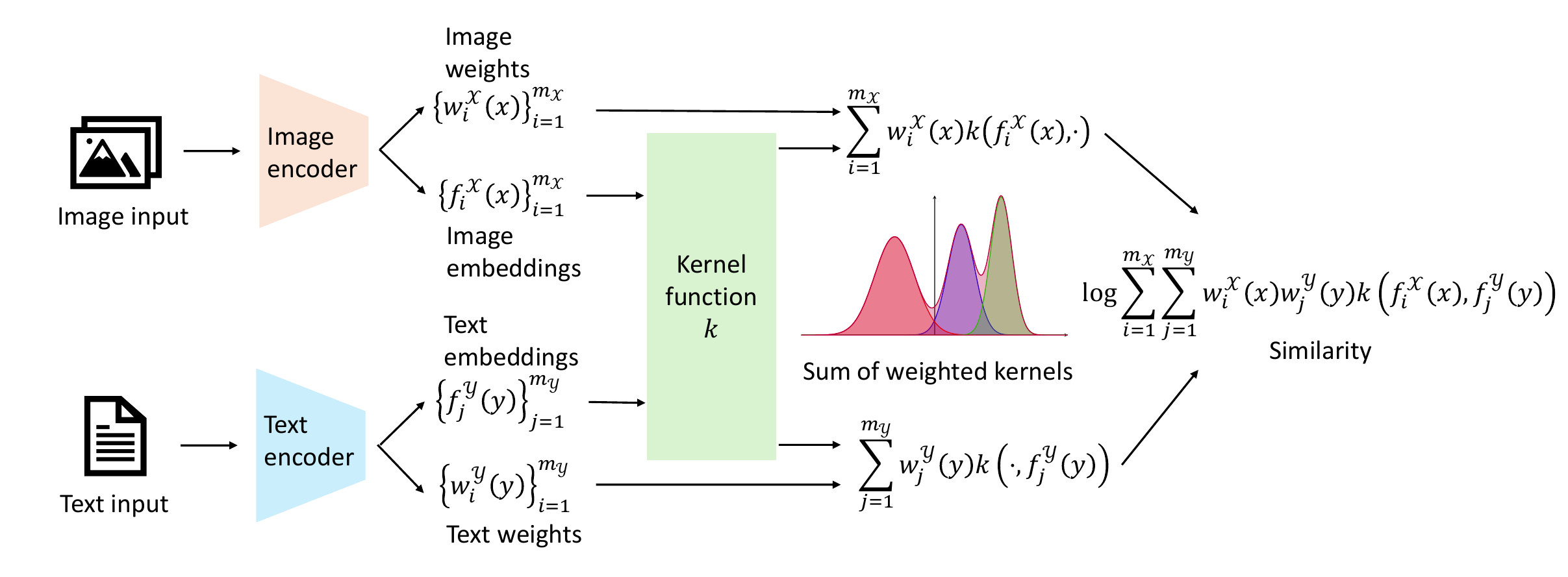}
\end{center}
\vspace{-12pt}
\caption{Overview of our proposed method. We project the image and text embedding into RKHS and calculate similarity by the logarithm of the inner product in this space.}
\label{fig: method}
\end{figure*}

Figure \ref{fig: method} shows a summary of our proposed method.
A comparable approach is implemented in WPSE \citep{uesaka2025wpse}; readers may refer to Figure 2 therein for a more detailed illustrative example.

Though our method is available by any positive kernel $k$, we primarily focus on the case of the Gaussian kernel. This is from two reasons. First is the capability of taking the value near zero. This is for approximating a situation where $p(x,y)=0$ holds. Such situations occur when the image and provided caption do not correspond at all, such as an image of a dog and a text \textit{this is a cat}. Second, with a Gaussian kernel, our method involves CLIP as a special case, as is stated later in Section \ref{section: comparison-clip}. This property provides a certain degree of assurance that our method will achieve performance at least comparable to that of CLIP.

\subsection{Comparison with other existing methods}\label{section: comparison}
In this section, we compare KME-CLIP with the two most relevant methods, CLIP and WPSE~\citep{uesaka2025wpse}.
For the full exposition of relevant literature, see \Cref{section: related work}.

\subsubsection{Comparison with CLIP}\label{section: comparison-clip}
% \paragraph{Comparison with CLIP.}
CLIP uses the form of $g^\mX(x)^\top g^\mY(y)/\tau$ as the similarity metric between the embedding vectors $g^\mX(x)$ and $g^\mY(y)$.
CLIP can be regarded as a special case of KME-CLIP,
when $g^\mX$ and $g^\mY$ are normalized vectors in $\R^d$.
Consider using the Gaussian kernel $k=k_\sigma$, \textit{i.e.}, $k_\sigma(u,v):=\exp(-\nor{u-v}^2 / 2\sigma^2)$.
If we let $m_\mX=m_\mY=1$, $w_1^{\mX}(x)=w_1^{\mY}(y)=1$, and $\sigma^2=\tau$,
then we have the following proposition:

\begin{proposition}\label{prop: kme_clip-clip}
    Under the setting above, the similarity function of KME-CLIP corresponds to that of CLIP excluding the addition of a constant.
\end{proposition}
%Then the similarity function of KME-CLIP corresponds to that of CLIP without scaling by a constant factor, since we apply $L^2$-normalization to the embedding vectors $g^\mX(x)$ and $g^\mY(y)$.
The proof is given in Appendix \ref{section: auxiliary}.
We remark that the additive constant in this statement does not affect the property of the model since it vanishes in the loss calculation.
From \Cref{prop: kme_clip-clip}, the similarity function of KME-CLIP is at least as expressive as CLIP if we use the same Euclidean embedding space $\R^d$.
It also represents a practical generalization of CLIP by incorporating learnable weights and utilizing larger size of point set in its implementation.
Moreover, while KME-CLIP apparently uses more embedding parameters than CLIP,
we can just reuse the discarded outputs of CLIP's transformer implementation,
thus the network size for obtaining CLIP and KME-CLIP is almost identical.

While CLIP can be viewed as a one-point version of our method, their theoretical properties differ fundamentally: our method leverages a linearity structure, whereas CLIP does not. The logarithmic term of our similarity function, $\ip{h_\theta^\mX(x), h_\theta^\mY(y)}_\mH$, exhibits a bi-linear structure with respect to functions of $x$ and $y$. In contrast, the logarithmic term of CLIP’s similarity function, $\exp(g^\mX(x)^\top g^\mY(y)/\tau)$, does not possess such a bi-linear structure due to the non-linear transformation introduced by the $\exp$ function.
\blue{This structural distinction is what Theorems \ref{thm: clip_limitation} and \ref{thm: kme-clip-superiority} make precise on a specific low-dimensional family of distributions.}

% \begin{figure}[htbp]
% \begin{center}
% \includegraphics[width=0.8\linewidth]{items/clip-kme-clip2.pdf}
% \end{center}
% \vspace{-12pt}
% \caption{Comparison of KME-CLIP and CLIP. CLIP can be regarded as one-point KME-CLIP.}
% \label{fig: clip-kme-clip}
% \end{figure}

\subsubsection{Comparison with WPSE}\label{section: comp-wpse}
% \paragraph{Comparison with WPSE.}
WPSE \citep{uesaka2025wpse} embeds datapoints into an RKHS in the same way as our method,
and it calculates similarity in the form of $\ip{u, v}_{\gH}/\tau$ without $\log$:
\begin{equation}
    S(x,y)=\frac1\tau\sum_{i,j} {w'_i}^{\mX}(x){w'_j}^{\mY}(y) k({f'_i}^{\mX}(x), {f'_j}^{\mY}(y)),
    \label{eq:wpse-kernel}
\end{equation}
where ${f'_i}^\mX$ and ${f'_i}^\mY$ are image and text encoders, ${w'_i}^\mX$ and ${w'_i}^\mY$ are image and text weights.
They also utilize multiple outputs from the Transformer architecture to obtain ${f'_i}^\mX$ and ${f'_i}^\mY$.
While the implementation methodology for generating embeddings and weights is identical to our approach,
the preferred choice of $k$ in WPSE is ``Gaussian + linear'' kernel $k(x, y) = \frac13 k_\sigma(x,y) + \frac23 x^\top y$
for better performance \citep{uesaka2025wpse},
unlike a single Gaussian in KME-CLIP.

However, the underlying motivation of WPSE is fundamentally different from KME-CLIP. WPSE is motivated by transforming CLIP embeddings into infinite-dimensional vectors from the perspective of representational capacity, whereas our approach aims at leveraging inherent linear structure in exp(PMI), which is ignored both by CLIP and WPSE. This distinction leads to the appearance of an additional $\log$ factor and results in differences in the theoretical and empirical properties discussed in the following remarks.
\begin{remark}[Theoretical difference]
\label{remark: wpse-theory}
Although \citet{uesaka2025wpse} also showed that WPSE can approximate PMI with small errors under the assumption that the PMI function is Lipschitz continuous, this assumption may be violated when the joint distribution contains a pair $(x, y)$ such that $p(x,y)=0$ while $p(x), p(y) > 0$, because $\log \frac{p(x,y)}{p(x)p(y)}$ diverges to $-\infty$.
In contrast, our theoretical analysis assumes Lipschitz continuity of $\exp(\text{PMI})$, rather than of PMI itself, as stated in \Cref{assumption: lipschitz}.
Consequently, our framework can accommodate cases in which PMI diverges to $-\infty$.
\end{remark}

\begin{remark}[Empirical difference]
    The additional ``log" in~\eqref{similarity_kernel} distinguishes the empirical behavior of KME-CLIP from that of WPSE reported later in Section~\ref{section: experiment}.
    Two empirical observations in \citet{uesaka2025wpse} are relevant to the comparison:
    \begin{itemize}
        \item \textbf{Kernel:} WPSE uses a mixture of nonlinear (Gaussian) and linear kernels as described below the definition in \eqref{eq:wpse-kernel}, since
        solely using a nonlinear or linear kernel deteriorates the performance \citep[Tables~6]{uesaka2025wpse}.
        \item \textbf{Weights:} When we restrict the weights in WPSE (with inverse multiquadric kernel) to be positive,
        then its zero-shot accuracy in CC3M (4\%) is far worse than 27\% of the case allowing negative weights \citep[Table~3]{uesaka2025wpse}.
    \end{itemize}
    \blue{Both observations concern configurations that differ from ours in more
    than the outer logarithm, and we are careful not to read them as evidence
    for the logarithm. Our matched control KME-no-log
    (\Cref{app:real-data-nolog}) uses a single Gaussian kernel with positive
    weights and a learned inverse length $b$, and it attains the highest
    $13$-dataset mean zero-shot accuracy among the four methods we compare.
    Positive weights with a single Gaussian are therefore viable without the
    outer logarithm, and what our experiments suggest is that the failure
    reported in \citet[Table~3]{uesaka2025wpse} is specific to that
    configuration (mixture of a linear kernel and an inverse multiquadric kernel with a fixed bandwidth) rather
    than a consequence of constraining the weights to be positive. The separate
    question of what the outer logarithm contributes is addressed in
    Section~\ref{sec: outer-log}.}
\end{remark}

\section{Theory}\label{section: theory}

\subsection{Theoretical guarantee for KME-CLIP}\label{section: theory_of_our_method}
In this section, we theoretically guarantee the performance of our method by proving the intuition given in Section~\ref{section: intuition}.

Specifically, by combining the following \Cref{thm: pmi_error_to_loss,thm: pmi_in_rkhs,thm: discrete_error_2},
we demonstrate that the population loss in \eqref{exp_of_loss} can be brought arbitrarily close to its theoretical optimal value using the KME-CLIP similarity function when the size of point set is sufficiently increased.
Furthermore, these results
can avoid the strong assumptions such as the Lipschitz continuity of PMI
as discussed in \Cref{section: comp-wpse},
thanks to the linearity of the inner product in RKHS.

\paragraph{Approximation of PMI implies near-optimal loss.}
First, we investigate how the difference between $\exp(\text{PMI})$ and $\exp(S(x,y))$ affects the deviation of the loss from its minimum value. The proof is given in \Cref{section: proof-thm3}.
\begin{theorem}\label{thm: pmi_error_to_loss}
    Let $0<\delta\le\epsilon\le 1/e^2$.
    Under \Cref{assumption: density},
    let a function $h:\mX\times\mY\to\R_+$ satisfy
    $h\ge\delta$ and $\lvert h(x,y)-\exp(\text{\rm PMI}(x,y))\rvert \le \epsilon$ for all $x\in\mX$ and $y\in\mY$.
    Then, we have
    \begin{equation}
        L_{\log h} \le \inf_{S}L_S + 3\sqrt{\epsilon} +
        \sqrt{\epsilon}\log\left(\frac1{\epsilon} + \frac1{\delta}\right),
        \label{eq: pmi_erro_to_loss}
    \end{equation}
    where $L_S$ is the population loss of CLIP for a similarity function
    $S$ defined in \eqref{exp_of_loss}.
\end{theorem}

This theorem states that if a function $h$ approximates $\exp(\mathrm{PMI})$ uniformly well, then the population loss $L_{\log h}$ is close to its minimum value.
Here, $h\ge \delta$ is a technical assumption,
but we can construct an example of $\delta=\epsilon$
from a function $\tilde{h}$ with $\lvert \tilde{h} - \exp(\text{PMI}(x, y))\rvert\le\epsilon$
by simply taking $h(x, y) = \max\{\epsilon, \tilde{h}(x, y)\}$.

\paragraph{Universal approximation of $\exp(\text{PMI})$ via kernel mean embeddings.}
Second, we show that $\exp(\text{PMI})$ can well be approximated by the integral form of the kernel mean embedding, using positive weight functions.
Before stating the result, we pose an additional assumption on the regularity of the conditional densities.
\begin{assumption}[Regularity of conditional densities]\label{assumption: lipschitz}
    We assume that there exists a universal constant $C_L>0$ such that,
    for every $x\in\mX$ and $y\in\mY$,
    $p(x|z)/p(x)$ and $p(y|z)/p(y)$ are $C_L$-Lipschitz continuous with respect to $z$.
\end{assumption}
With this assumption, we have the following result. We defer its proof to \Cref{section: proof-thm4}.

\begin{theorem}\label{thm: pmi_in_rkhs}
Suppose that Assumptions~\ref{assumption: density}--\ref{assumption: lipschitz} hold.
Let $k$ be a Gaussian kernel on $\R^d$ with length-scale $\sigma$,
i.e., $k(u, v)=\exp(-\lVert u-v\rVert^2/(2\sigma^2))$.
For an arbitrary $\epsilon>0$, there exist an appropriate choice of $\sigma$,
a probability measure $\mu$ on $\R^d$,
and \textbf{positive} functions $g_X^x, g_Y^y\in L^2(\mu)$ for all $x\in\mX$ and $y\in\mY$
such that the following holds: For each $x\in\mX$ and $y\in\mY$, we have
\begin{align*}
    \biggl\lvert
    \ip{\int_{\R^d}k(\cdot, u)g_X^x(u)\dd\mu(u),
        \int_{\R^d}k(\cdot, u)g_Y^y(u)\dd\mu(u)
    }_{\mathcal{H}}-
    \exp(\text{\rm PMI}(x, y))
    \biggr\rvert
    < \epsilon.
\end{align*}
\end{theorem}
While there is no constraint on $\mu$,
from the proof,
we can additionally take $\mu$ to be a measure over the unit sphere of $\R^d$ (if $d\ge2$),
following the convention of CLIP.

\paragraph{Discretization error of kernel mean embeddings.}
Finally, we show the following discretization result since it is difficult to compute the integral form of kernel mean embeddings in a practical implementation.
While it essentially follows from the previous results on positively-weighted kernel quadrature \citep{chen2010super,bach2012equivalence,hayakawa2022positively},
we prove it in \Cref{section: proof-thm5} for completeness.

\begin{theorem}\label{thm: discrete_error_2}
    % Let us assume that $\int_{\gZ}k(z,z)g(z)^2\dd\mu(z), \iint_{\gZ\times\gZ}k(z,z')g(z)g(z')\dd\mu(z)\dd\mu(z')<\infty$ and denote $c:=\int_{\gZ}k(z,z)g(z)^2\dd\mu(z)-\iint_{\gZ\times\gZ}k(z,z')g(z)g(z')\dd\mu(z)\dd\mu(z')$.
    Let $k$ be a positive definite kernel on $\R^d$ such that $K:=\sup_{u\in\R^d}k(u, u)<\infty$.
    Given a probability measure $\mu$ over $\R^d$
    and $g\in L^2(\mu)$,
    for each positive integer $m$,
    there exists $u_1,...,u_{m}\in\R^d$ such that
    \begin{align*}
        \nor{
            \int_{\R^d} k(u, \cdot)g(u)\dd\mu(u)
            - \frac1m\sum_{i=1}^{m} k(u_i, \cdot)g(u_i)
        }_{\gH}
        \le \frac{\sqrt{K}\lVert g\rVert_{L^2(\mu)}}{\sqrt{m}}&,
    \end{align*}
    where $\lVert\cdot\rVert_\gH$ denotes the norm in $\gH$.
\end{theorem}

As this theorem states, the approximation error of the discretization decays by $O(1/\sqrt{m})$,
where $m$ is the size of the point set used in the embedding (see, e.g., \eqref{eq: kme-embedding}).
Note that we can take $K=1$ as long as we use Gaussian kernels.

\subsection{Theoretical comparison of CLIP and KME-CLIP under low modality gap}
\label{section: theory_clip}
In this section, we introduce a toy example that illustrates a scenario where CLIP encounters significant difficulty in approximating PMI through its similarity function, while KME-CLIP does not.
Although this represents a simplified model, it effectively demonstrates \blue{a structural limitation} of CLIP and the inherent advantages of our proposed method.

\paragraph{Setup.}
Let us naturally define in-modal similarities defined by CLIP and KME-CLIP:
for $x,x'\in\mX$, we define
$
    S^\mX(x,x')=g^\mX(x)^\top g^\mX(x')/\tau
$
when using CLIP and 
$
    S^\mX(x,x')=\log\ip{h_\theta^\mX(x), h_\theta^\mX(x')}_\mH
$
when the model is KME-CLIP.
For readability, under \Cref{assumption: conditional},
we specifically write $p(x)$ as $p_X(x)$ and $p(x|z)$ as $p_{X|Z}(x|z)$ in this section (same for $Y$).
Let us also define $p_{X}(x,x'):=\int p_{X|Z}(x|z)p_{X|Z}(x'|z)\dd\rho(z)$.
With these notations, we can also define the in-modal PMI
as
$$
    \text{PMI}^\mX(x,x'):=\log \frac{p_{X}(x,x')}{p_X(x)p_X(x')}.
$$

Now we introduce an interpretation of a model with low modality gap:
\begin{definition}[Low modality gap]\label{definition: low-modality-gap}
    There exists $\delta>0$ such that the similarities $S$ and $S^\mX$ satisfy,
    for any $x,x'\in\mX$ with $\text{\rm PMI}^\mX(x,x')<\log\delta$,
    there exists $y\in\mY$ such that $S^\mX(x,x')\le S(x, y)$.
\end{definition}
This property keeps the model (either CLIP or KME-CLIP) from embedding different modalities 
to regions completely separated to each other, preventing a large modality gap as discussed by \citet{modality-gap}. Low modality gap is a desirable property for embeddings since large modality gap leads to a degradation in model performance.
Incorporating
in-modal self-supervised learning like SimCLR \citep{chen2020simple} into multimodal representation learning is
also done by \citet{mu2022slip}.

Next, let us introduce the data-generation process on which we compare CLIP and KME-CLIP.
\begin{assumption}[$2$-mixture model]\label{assumption: 2-mixture}
    Under Assumptions~\ref{assumption: density} and \ref{assumption: conditional},
    the latent and state spaces are given by $\gZ=\{1,\ldots, N\}$ and
    $\mX=\mY=\{(i,j)\mid i,j\in\gZ\}$
    with \textbf{unordered pairing}, i.e., $(i, j)=(j, i)$.
    The distribution of $Z$, $\rho$ is given by the uniform distribution over $\gZ$,
    and for each $i,j,\ell\in\gZ$, we let
    $$
        p_{X|Z}((i, j)\mid \ell) = p_{Y|Z}((i, j)\mid \ell) = \frac1{N+1}(\delta^{i\ell}+\delta^{j\ell}),
    $$
    where $\delta^{i\ell}$ and $\delta^{j\ell}$ are Kronecker's delta.
\end{assumption}
This setting represents the situation like this:
There exist classes $1,\ldots,N$ and for each class $i$, there exists a datapoint $(i, i)$ representing purely this class. 
The other datapoints, $(i, j)$ for $i\ne j$, represent a half-and-half mixture of classes $i$ and $j$.

\paragraph{Results.}
Under the setting above, the following theorem states that
CLIP has a difficulty in expressing $\exp(\text{PMI})$ when $N$ is large,
even if we allow the degree of freedom regarding the constant term in \Cref{thm: pmi-minimzer}. We defer its proof to \Cref{section: proof-thm6}.
\begin{theorem}[\blue{Low-dimensional separation:} limitation of CLIP in PMI approximation]\label{thm: clip_limitation}
    Under Assumption~\ref{assumption: 2-mixture},
    let us also assume $N>9^d$, where the CLIP embeddings $g^\mX(x)$, $g^\mY(y)$
    are on the unit sphere of $\R^d$.
    Then, at least one of the following holds:
    (a) CLIP embeddings do not satisfy the low modality gap condition.
    (b) For any choice of constants $\alpha,\tau>0$, there exist $x\in\mX$ and $y\in\mY$ such that
    \[
        \left|\alpha\exp\left(g^\mX(x)^\top g^\mY(y)/\tau\right)-\exp(\text{\rm PMI}(x, y))\right| \ge N/4.
    \]
    
\end{theorem}
\blue{We emphasize that \Cref{thm: clip_limitation} is a low-dimensional toy
separation rather than a statement about our experiments.
The hypothesis
$N>9^d$ is not satisfied in our experiments at the embedding dimension
$d=512$ and would require an astronomically large $N$ at that dimension,
so the theorem places no bound on the downstream performance of CLIP in
our real-data settings.
Its role is to
exhibit a concrete distribution on which the non-bilinear form of CLIP's
similarity provably cannot represent $\exp(\mathrm{PMI})$ together with a low
modality gap, and thereby to motivate the bilinear RKHS construction.}

In contrast \blue{to the lower bound for CLIP in Theorem~\ref{thm: clip_limitation}}, our method can exploit the sparse structure of the data distribution as follows.
We give its proof in \Cref{section: proof-thm7}.
\begin{theorem}[PMI approximation by KME-CLIP under low modality gap]\label{thm: kme-clip-superiority}
    Under the same setting as \Cref{thm: clip_limitation},
    let us use Gaussian kernel on $\R^d$ ($d\ge2$) for KME-CLIP with $m_\mX=m_\mY=2$.
    For any $\epsilon>0$, there exists an appropriate choice of the length-scale $\sigma$,
    encoders, and weight functions that satisfy the following conditions at the same time:
    (a) These KME-CLIP embeddings satisfy a low modality gap condition.
    (b)~For each $x\in\mX$ and $y\in\mY$,
    and the embeddings $h_\theta^\mX$ and $h_\theta^\mY$ given by \eqref{eq: kme-embedding}, we have
        \begin{align*}
            &\left\lvert
            \ip{h_\theta^\mX(x), h_\theta^\mY(y)}_\gH-
            \exp(\text{\rm PMI}(x, y))\right\rvert<\epsilon.
        \end{align*}
\end{theorem}

In Section~\ref{sec: experiment-gap}, we provide empirical diagnostics related
to modality gap and true-pair alignment, as supporting evidence for this
low-modality-gap perspective.

\blue{A continuous extension of the $2$-mixture model in this section also
underlies the known-PMI experiments of \Cref{app:known-pmi}, where
$\exp(\operatorname{PMI})$ is available in closed form, so that the
approximation quality of each similarity can be measured directly rather than
through downstream accuracy.}
\section{Experiments}\label{section: experiment}
In this section, we empirically demonstrate the effectiveness of our method across several downstream tasks.
Below, we describe the outline of the experimental setup.

%\subsection{Pre-training}\label{section: ex_pretrain}
% \textbf{Dataset} \\
For pre-training, we utilized the training splits of Conceptual Captions 3M (CC3M) \citep{sharma-etal-2018-conceptual} and Conceptual Captions 12M (CC12M) \citep{changpinyo2021conceptual} datasets.
We used the same dataset for training CLIP, WPSE, and KME-CLIP for a fair comparison.
We adopted a non-pretrained ViT-B/16 encoder \citep{dosovitskiy2020image} as the base architecture for the image encoder and a non-pretrained Transformer encoder \citep{vaswani2017attention} with width $512$, $12$ layers, and $8$ attention heads for the text encoder across CLIP, WPSE, and KME-CLIP.
For optimization, we employed AdamW with a cosine learning rate scheduler incorporating linear warm-up, using a batch size of $1024$. The hyperparameter configurations followed those established in \cite{uesaka2025wpse}, with consistent settings maintained across CLIP, WPSE, and KME-CLIP implementations to ensure fair comparison.

Note that our CLIP and WPSE baselines are retrained under this shared protocol rather than transcribed from \cite{uesaka2025wpse}, and their absolute numbers differ from those reported in the original paper.
One concrete difference is the global batch size: \cite{uesaka2025wpse} use $2048$, whereas we use $1024$ here and $512$ for the three-seed study in \Cref{app:real-data-nolog} to fit our compute budget. Since the batch size changes the difficulty of the in-batch contrastive objective, absolute numbers are not comparable across papers, and our claims rest on the within-paper comparisons.

As for the implementation of KME-CLIP, we employed the Gaussian kernel for the non-linear kernel function $k$, \textit{i.e.}, $k(u,v):=\exp(-\nor{u-v}^2/2\sigma^2)$. 
To avoid numerical instability when $\sigma$ approaches zero, we reparameterized \blue{by learning an unconstrained scalar $b$ and using $b^2=1/\sigma^2$ in the kernel}.
We adopted a softplus function, \textit{i.e.}, $\log(1+e^{x})$ for the activation function for the positivity.
For implementation details regarding CLIP, WPSE and KME-CLIP as well as comprehensive hyperparameter specifications, refer to Appendix \ref{appendix: ex_additional}.

We evaluate performance on retrieval, zero-shot classification, and linear classification—all standard downstream tasks for assessing multimodal representation learning capabilities.
We defer detailed information regarding linear classification to \Cref{section: linear_classification}.

\blue{
\paragraph{Summary of findings.}
We summarize the outcome of all experiments here, including those deferred to the appendix.
\begin{itemize}
    \setlength{\itemsep}{0pt}
    \item \textbf{Retrieval and zero-shot classification} (Tables~\ref{table: retrieval} and \ref{table: zero-shot}): KME-CLIP attains the highest average in both, though not in every cell.
    \item \textbf{Point-set size and cost} (\Cref{section: ablation}, \Cref{section: computational_cost}): accuracy degrades gracefully as the point set shrinks, and at $m_\mX=10$ KME-CLIP matches the GPU count and training time of CLIP.
    \item \textbf{Linear classification} (\Cref{section: linear_classification}): KME-CLIP is comparable to the baselines under CC3M pretraining but the weakest of the three under CC12M, so we do not claim that it yields better linearly separable features.
    \item \textbf{Modality gap} (\Cref{sec: experiment-gap}): KME-CLIP has a smaller vector-based gap than CLIP on all three datasets, while true-pair alignment is mixed.
    \item \textbf{Role of the outer logarithm} (\Cref{sec: outer-log}): a matched control without the logarithm (KME-no-log) is stronger on CC3M downstream metrics, while the logarithm gives the more faithful PMI approximation in the known-PMI setting.
\end{itemize}
}

\subsection{Retrieval}
We evaluated the retrieval ability of the embeddings, by the validation splits of CC3M, MSCOCO \citep{cococaption}, and Flickr30K \citep{flikr}. The choice of dataset is following previous works like \cite{yao2022filip} or \cite{desai2023hyperbolic}.
The (image-to-)text and (text-to-)image retrieval tasks we employ here
are the primary target of minimizing CLIP-like losses.

We show the result in Table \ref{table: retrieval}. 
% \blue{KME-CLIP has the highest reported average retrieval performance in
% these single runs, although it does not outperform both baselines in every
% cell.}
% \blue{Downstream retrieval does not directly measure PMI fidelity; we evaluate
% it in a synthetic setting with analytically known PMI in
% Appendix~\ref{app:known-pmi}.}
These numbers are single runs at batch size $1024$; three-seed results under a matched protocol on CC3M, including standard deviations, are reported in \Cref{app:real-data-nolog}. KME-CLIP has the
highest reported average, although it does not outperform both baselines in
every cell: CLIP is better on MSCOCO text retrieval under CC12M pretraining.

% \begin{table}[htbp]
% \caption{Retrieval results. We report the top-1 accuracy of image-to-text retrieval and text-to-image retrieval separately.}
% \label{table: retrieval}
% %\vskip 0.15in
% \begin{center}
% \begin{scriptsize}
% \tabcolsep = 3pt
% \begin{tabular}{@{}llrrrp{.15cm}rrrp{.15cm}rrr@{}}%%\toprule
%      &      & \multicolumn{3}{c}{Text Retrieval} && \multicolumn{3}{c}{Image Retrieval}  \\
% \cmidrule{3-5} \cmidrule{7-9}
%      &   Model   & CC3M   & MSCOCO   & Flickr30K  && CC3M   & MSCOCO   & Flickr30K  \\ \midrule
% \multirow{3}{*}{CC3M} & CLIP  & 21.97 & 13.64 & 26.46 && 25.37 & 14.57 & 27.68 \\
%      & WPSE  & 21.23 & \textbf{14.70} & 26.80 && 25.36 & 15.39 & 28.40 \\
%      & KME-CLIP & \textbf{24.22} & 14.13 & \textbf{28.18} && \textbf{26.91} & \textbf{15.74} & \textbf{30.52} \\ \midrule
% \multirow{3}{*}{CC12M} & CLIP  & 22.85 & \textbf{24.57} & 44.78 && 22.72 & 24.74 & 46.06 \\
%      & WPSE  & 22.44 & 24.32 & 45.08 && \textbf{23.71} & 24.55 & 47.08 \\
%      & KME-CLIP & \textbf{23.39} & 23.54 & \textbf{46.80} && 23.70 & \textbf{24.93} & \textbf{47.92} \\ 
% \bottomrule
% \end{tabular}
% \end{scriptsize}
% \end{center}
% %\vskip -0.1in
% \vspace{-10pt}
% \end{table}

% テーブル1: Text Retrieval
\begin{table*}[htbp]
\caption{Text retrieval (top) and image retrieval (bottom) results. We report the top-1, 5, 10 accuracy (\%).}
\label{table: retrieval}
\begin{center}
\begin{scriptsize}
\begin{tabular}{@{}llrrrrrrrrrr@{}}
\toprule
     &      & \multicolumn{3}{c}{CC3M} & \multicolumn{3}{c}{MSCOCO} & \multicolumn{3}{c}{Flickr30K} \\
\cmidrule(lr){3-5} \cmidrule(lr){6-8} \cmidrule(lr){9-11}
     Text retrieval &     Model      & Top1 & Top5 & Top10 & Top1 & Top5 & Top10 & Top1 & Top5 & Top10 \\ 
\midrule
\multirow{3}{*}{CC3M} 
     & CLIP      & 21.95 & 41.97 & 51.01 & 13.64 & 32.34 & 43.57 & 26.46 & 52.18 & 62.68 \\
     & WPSE      & 21.39 & 41.36 & 50.51 & \textbf{14.78} & 33.75 & 44.86 & 27.22 & 52.80 & 63.92 \\
     & KME-CLIP  & \textbf{24.22} & \textbf{44.85} & \textbf{53.51} & 14.14 & \textbf{34.25} & \textbf{45.65} & \textbf{28.18} & \textbf{55.72} & \textbf{65.92} \\ 
\midrule
\multirow{3}{*}{CC12M} 
     & CLIP      & 22.82 & 42.89 & 52.22 & \textbf{24.56} & 48.25 & 60.15 & 44.76 & 73.30 & 81.94 \\
     & WPSE      & 22.59 & 42.70 & 51.94 & 24.29 & 48.85 & 61.07 & 45.80 & 73.74 & 82.40 \\
     & KME-CLIP  & \textbf{23.36} & \textbf{44.38} & \textbf{54.19} & 23.54 & \textbf{48.87} & \textbf{61.24} & \textbf{46.80} & \textbf{75.90} & \textbf{84.68} \\ 
\bottomrule
\vspace{0cm}
\end{tabular}
\begin{tabular}{@{}llrrrrrrrrrr@{}}
\toprule
     &      & \multicolumn{3}{c}{CC3M} & \multicolumn{3}{c}{MSCOCO} & \multicolumn{3}{c}{Flickr30K} \\
\cmidrule(lr){3-5} \cmidrule(lr){6-8} \cmidrule(lr){9-11}
     Image retrieval &      Model     & Top1 & Top5 & Top10 & Top1 & Top5 & Top10 & Top1 & Top5 & Top10 \\ 
\midrule
\multirow{3}{*}{CC3M} 
     & CLIP      & 25.36 & 46.01 & 54.91 & 14.57 & 34.52 & 45.29 & 27.68 & 53.24 & 64.42 \\
     & WPSE      & 25.36 & 46.63 & 56.05 & 15.36 & 35.29 & 46.48 & 28.48 & 54.14 & 64.64 \\
     & KME-CLIP  & \textbf{26.90} & \textbf{49.05} & \textbf{58.09} & \textbf{15.74} & \textbf{35.94} & \textbf{47.25} & \textbf{30.52} & \textbf{57.56} & \textbf{68.42} \\ 
\midrule
\multirow{3}{*}{CC12M} 
     & CLIP      & 22.71 & 43.37 & 52.55 & 24.74 & 48.93 & 60.36 & 46.10 & 73.92 & 82.24 \\
     & WPSE      & \textbf{23.76} & 44.33 & 53.85 & 24.73 & 49.45 & 61.46 & 47.12 & 74.96 & 83.04 \\
     & KME-CLIP  & 23.70 & \textbf{45.51} & \textbf{55.14} & \textbf{24.93} & \textbf{50.47} & \textbf{62.53} & \textbf{47.92} & \textbf{76.82} & \textbf{85.16} \\ 
\bottomrule
\end{tabular}

\end{scriptsize}
\end{center}
\vspace{-10pt}
\end{table*}

\subsection{Zero-shot classification}
Zero-shot classification is an approach for image classification where, given a list of class names, the method classifies images by performing image-to-text retrieval using captions generated from these class names.
We evaluated it by the test splits (for datasets without a test split, the validation split was used) of the following 13 benchmark datasets: ImageNet \citep{russakovsky2015imagenet}, CIFAR-10 \citep{krizhevsky2009learning}, CIFAR-100 \citep{krizhevsky2009learning},
STL-10 \citep{coates2011analysis}, Food-101 \citep{bossard2014food}, Caltech-101 \citep{fei2006one}, Stanford Cars \citep{krause20133d},
FGVC Aircraft \citep{maji2013fine}, Oxford Flowers \citep{nilsback2008automated}, EuroSAT \citep{helber2019eurosat}, Describable Textures Dataset (DTD) \citep{cimpoi2014describing}, Oxford Pets \citep{parkhi2012cats}, and SUN397 \citep{xiao2010sun}.
Following SLIP \citep{mu2022slip}, we adopted prompt ensembling and utilized prompts provided by SLIP for each dataset.

We show the results in Table~\ref{table: zero-shot}.
These come from the same single runs at batch size
$1024$ as in the previous section, and the gains are not uniform across the individual datasets; the corresponding three-seed zero-shot results are given in \Cref{app:real-data-nolog}.

\begin{table*}[htbp] %[htbp]
\caption{Zero-shot classification results. Mean per-class accuracy (\%) is reported for Caltech-101, Aircraft, Flowers, and Pets datasets. Top-1 accuracy (\%) is used for all other datasets.
%Models were trained on CC3M and CC12M.
%Gray highlighting indicates the models based on weighted point clouds.
}
\label{table: zero-shot}
%\vskip 0.15in
\begin{center}
\begin{scriptsize}
\tabcolsep = 3pt
\begin{tabular}{cccccccccccccccc}
\toprule
%\multicolumn{2}{c}{Model}
 & Model & \rotatebox{90}{Average} &
\rotatebox{90}{ImageNet} & \rotatebox{90}{\tiny CIFAR-10} & \rotatebox{90}{\tiny CIFAR-100} & \rotatebox{90}{STL-10} & \rotatebox{90}{Food-101} & \rotatebox{90}{\tiny Caltech-101} & \rotatebox{90}{Cars} & \rotatebox{90}{Aircraft} &
\rotatebox{90}{Flowers} & \rotatebox{90}{EuroSAT} &
\rotatebox{90}{DTD} & \rotatebox{90}{Pets} &
\rotatebox{90}{SUN397} \\
\midrule
\multirow{3}{*}{CC3M} 
& CLIP & 26.33 & 20.95 & 56.77 & 24.38 & 82.80 & 14.06 & 50.07 & \textbf{1.46} & \textbf{1.20} & 11.64 & 13.84 & \textbf{13.88} & 14.10 & 37.20 \\
& WPSE & 27.38 & 22.17 & 59.40 & 30.78 & 81.64 & 14.80 & 50.70 & 1.27 & 1.13 & 14.01 & 16.20 & 13.83 & 12.78 & 37.27 \\
&  KME-CLIP & \textbf{29.31} & \textbf{22.66} & \textbf{64.31} & \textbf{31.32} & \textbf{85.23} & \textbf{16.17} & \textbf{53.67} & 1.29 & 1.04 & \textbf{16.54} & \textbf{17.68} & 12.39 & \textbf{16.26} & \textbf{42.49} \\
\midrule
\multirow{3}{*}{CC12M}
& CLIP & 43.48 & 36.99 & 75.41 & 42.77 & 92.36 & 45.97 & 71.60 & 17.44 & 2.48 & \textbf{25.58} & 31.72 & 19.47 & 55.59 & 47.88  \\
& WPSE & 43.37 & 36.92 & 74.54 & 42.22 & 92.24 & 43.30 & 71.53 & \textbf{18.62} & \textbf{3.80} & 22.32 & \textbf{32.98} & 19.47 & 54.52 & 51.36 \\
&  KME-CLIP & \textbf{45.67} & \textbf{39.07} & \textbf{78.06} & \textbf{46.63} & \textbf{92.74} & \textbf{49.14} & \textbf{76.62} & 17.52 & 3.45 & 25.00 & 31.28 & \textbf{21.60} & \textbf{60.67} & \textbf{51.94}  \\
\bottomrule
\end{tabular}
\end{scriptsize}
\end{center}
%\vskip -0.1in
\vspace{-10pt}
\end{table*}

\subsection{Ablation study: Reducing the size of point set}\label{section: ablation}
We investigated the impact of the size of point set $m_\mX$ in KME-CLIP on retrieval performance using the validation split of CC3M.
To reduce the point set size $m_\mX$ in the image encoder, we utilized only a subset of tokens from the beginning of the $197$ outputs generated by the ViT architecture.
We maintained the original point set size $m_\mY=77$ for the text encoder without modification.
All other training settings followed the configuration described earlier in this section.

\begin{table*}[h]
    \caption{The average of top1 accuracy (\%) for text retrieval and image retrieval tasks.}
    \label{table: reduce_point}
    \begin{center}
    \begin{tabular}{lrrrrrr}
        \toprule
       Image point-set size $m_\mX$ & (CLIP) & 2 & 10 & 50 & 100 & 197 \\ \midrule
       Top-1 accuracy & (23.66) & 23.71 & 23.93 & 24.35 & 24.79 & 25.57 \\ \bottomrule
    \end{tabular}
    \end{center}
    \vskip -0.2in
\end{table*}

% \begin{table}[htbp]
%   \centering
%   \caption{Accuracy vs. Size of Point Cloud}
%   \label{tab:pointcloud_accuracy}
%   \begin{tabular}{l*{7}{S[table-format=2.2]}}
%     \toprule
%     & \multicolumn{7}{c}{Size of Point Cloud} \\
%     \cmidrule(lr){2-8}
%     & {1024} & {2048} & {4096} & {8192} & {16384} & {32768} & {65536} \\
%     \midrule
%     Accuracy (\%) & 85.67 & 87.32 & 89.45 & 91.23 & 92.78 & 93.56 & 94.12 \\
%     \bottomrule
%   \end{tabular}
% \end{table}

We report the top-1 accuracy in Table \ref{table: reduce_point}.
While the performance of KME-CLIP decreases as the point set size is reduced, it remains above CLIP at every tested size, although at $m_\mX=2$ and $m_\mX=10$ the margins ($0.05$ and $0.27$ points) are smaller than the seed-to-seed variation observed in \Cref{app:real-data-nolog} and we do not read them as meaningful.
See \Cref{section: computational_cost} for a more detailed comparison including their computational cost.
This result also indicates that larger size of point set leads to high performance of KME-CLIP, as is consistent with the theoretical result in Theorem \ref{thm: discrete_error_2}.

\subsection{Modality-gap diagnostics}
\label{sec: experiment-gap}

\begin{table}[htbp]
\centering
\caption{Modality-gap and true-pair alignment diagnostics. Gap is computed from vector embeddings, and lower is better. Align is computed by cosine similarity for CLIP and barycentric KME-CLIP, and by the constant-corrected LogSim diagnostic for KME-CLIP; higher is better.}
\label{tab:modality_gap}
\small
\begin{tabular}{llccc}
\toprule
Dataset & Metric & CLIP & KME-CLIP (Bary) & KME-CLIP (LogSim) \\
\midrule
\multirow{2}{*}{MSCOCO}
& Gap ($\downarrow$) & 0.508 & 0.402 & -- \\
& Align ($\uparrow$) & 0.380 & 0.418 & 0.494 \\
\midrule
\multirow{2}{*}{ImageNet}
& Gap ($\downarrow$) & 0.426 & 0.357 & -- \\
& Align ($\uparrow$) & 0.298 & 0.318 & 0.408 \\
\midrule
\multirow{2}{*}{CIFAR-10}
& Gap ($\downarrow$) & 0.826 & 0.754 & -- \\
& Align ($\uparrow$) & 0.300 & 0.278 & 0.432 \\
\bottomrule
\end{tabular}
\end{table}
In Section~\ref{section: theory_clip}, our theory suggests that KME-CLIP can
improve PMI approximation without requiring an unfavorable modality gap. The
preceding experiments evaluate the accuracy side of this claim, but they do not
show whether the learned representations preserve favorable cross-modal geometry.
We therefore complement the accuracy evaluations with modality-gap diagnostics given in this section,
using them as empirical support for the low-modality-gap perspective rather than
as a standalone claim that KME-CLIP solves modality gap in general.

We evaluate CLIP and KME-CLIP using two modality-gap diagnostics: the
modality gap and true-pair alignment. For a pair of vector embeddings
$a^\mX:\mX\to\R^d$ and $a^\mY:\mY\to\R^d$, we define
\[
    \mathrm{Gap}(a^\mX,a^\mY)
    :=
    \left\lVert
        \E_{p(x)}[a^\mX(x)]-\E_{p(y)}[a^\mY(y)]
    \right\rVert,\qquad 
    \mathrm{Align}(a^\mX,a^\mY)
    :=
    \E_{p(x,y)}\!\left[
        a^\mX(x)^\top a^\mY(y)
    \right],
\]
The first quantity is the conventional modality gap, where lower is better
\citep{modality-gap,grassucci2026closing}. The second quantity measures
the average cosine similarity of matched image--text pairs, where higher is
better \citep{grassucci2026closing}.
For CLIP, we directly use the normalized CLIP embeddings $g^\mX$ and $g^\mY$,
so that
    $
        \mathrm{Gap}_{\mathrm{CLIP}}
        :=
        \mathrm{Gap}(g^\mX,g^\mY)
        $ and $
        \mathrm{Align}_{\mathrm{CLIP}}
        :=
        \mathrm{Align}(g^\mX,g^\mY).
    $
For KME-CLIP, since it outputs weighted point sets rather than single vector
embeddings, we compute the diagnostics in the following two ways.
\begin{itemize}
    \item \textbf{Bary.}
    For KME-CLIP, we first form barycentric vector embeddings by taking the
    learned positive-weighted average of token embeddings and normalizing the
    result:
    \[
        \bar g^\mX(x)
        :=
        \frac{
            \sum_{i=1}^{m_\mX} w_i^\mX(x) f_i^\mX(x)
        }{
            \left\|
                \sum_{i=1}^{m_\mX} w_i^\mX(x) f_i^\mX(x)
            \right\|
        },
        \qquad
        \bar g^\mY(y)
        :=
        \frac{
            \sum_{j=1}^{m_\mY} w_j^\mY(y) f_j^\mY(y)
        }{
            \left\|
                \sum_{j=1}^{m_\mY} w_j^\mY(y) f_j^\mY(y)
            \right\|
        }.
    \]
    We then compute
    $
        \mathrm{Gap}_{\mathrm{Bary}}
        :=
        \mathrm{Gap}(\bar g^\mX,\bar g^\mY)
        $ and $
        \mathrm{Align}_{\mathrm{Bary}}
        :=
        \mathrm{Align}(\bar g^\mX,\bar g^\mY).
    $
    This gives a direct vector-space comparison with CLIP, but it ignores the
    kernel used in the actual KME-CLIP similarity.

    \item \textbf{LogSim.}
    We also report a kernel-aware true-pair alignment diagnostic, denoted by
    LogSim. We use a function $\mathrm{LogSim}_\theta:\mX\times\mY\to\R$
    given by a constant-corrected log-similarity of
    KME-CLIP (see Appendix~\ref{app: logsim} for a precise definition).
    It is defined so that, in the one-point Gaussian-kernel
    special case of Proposition~\ref{prop: kme_clip-clip}, it reduces exactly
    to the usual CLIP cosine score. We report
    \[
        \mathrm{Align}_{\mathrm{LogSim}}
        :=
        \E_{p(x,y)}[\mathrm{LogSim}_\theta(x,y)].
    \]
    Since it is
    not defined through a vector embedding, we do not compute a corresponding Gap value.
    \blue{Unlike the two conventional metrics, LogSim is a diagnostic we define
    here, and it depends on the kernel geometry of KME-CLIP rather than on a shared
    vector space. We therefore read the barycentric columns as the neutral
    comparison and LogSim as a kernel-aware supplement.}
\end{itemize}

The results are shown in Table~\ref{tab:modality_gap}. The barycentric
representation of KME-CLIP yields a smaller vector-based Gap than CLIP on all
evaluated datasets. Pair alignment is more nuanced: the barycentric score
improves on MSCOCO and ImageNet but not on CIFAR-10, while LogSim gives higher
kernel-aware alignment scores than the CLIP baseline on all three datasets.

\blue{
\subsection{The role of the outer logarithm}
\label{sec: outer-log}

Our derivation places the logarithm outside the RKHS inner product: the inner
product models $\exp(\operatorname{PMI})$, which is the object with linear
structure, and the logarithm recovers PMI itself as the similarity. Relative to
CLIP, however, KME-CLIP changes two things at once. It replaces a single global
image--text dot product with a weighted interaction over all image and text
token pairs, and it takes the logarithm of that interaction. The first change
by itself makes the similarity more expressive: a token-level late interaction
can improve retrieval simply because different words match different image
regions \citep{yao2022filip}, independently of the PMI interpretation. A
comparison against CLIP therefore cannot separate the two effects.

We address this from two sides. Analytically,
\Cref{prop: kme_clip-clip} identifies CLIP with the one-point special case of
KME-CLIP, which locates the two similarity classes relative to each other;
our CLIP baseline is trained with its own standard implementation rather than
as a matched one-point run, so we do not treat it as a controlled ablation of
the token-level interaction. Empirically, we introduce a matched control,
KME-no-log, which keeps that interaction and replaces the similarity
$S_{\mathrm{KME}}(x,y)=\log\langle h_\theta^\mX(x),h_\theta^\mY(y)\rangle_{\gH}$ with
$S_{\mathrm{nolog}}(x,y):=\langle h_\theta^\mX(x),h_\theta^\mY(y)\rangle_{\gH}$,
and is otherwise identical in parameterization: the same encoders and point-set
sizes, the same softplus-parameterized positive weights, the same single
Gaussian kernel with learned bandwidth, and the same initialization and
optimizer. The two models are trained separately and therefore produce
different point sets; what is matched is the construction, so the outer
logarithm is the only design difference between them.

The two evaluations do not agree. On CC3M with three seeds at batch size $512$,
KME-no-log has the higher mean in all $18$ retrieval entries and the higher
$13$-dataset mean zero-shot accuracy ($29.18\pm0.17$ against $28.68\pm0.96$;
\Cref{app:real-data-nolog}). In the synthetic setting where
$\exp(\operatorname{PMI})$ is known analytically, the ordering is reversed:
KME-CLIP attains the lowest conditional KL and the lowest relative $L^2$ error
among the four methods at embedding dimension $d\in\{8,16\}$ for every tested data setting, and it reaches
$98.4\%$--$99.8\%$ of the analytic Bayes ranker's top-10 accuracy at $d\ge4$,
where KME-no-log saturates between $89.3\%$ and $94.4\%$
(\Cref{app:known-pmi}).

It is worth being precise about what the logarithm can and cannot do. For a
fixed set of weights and point sets, the logarithm is strictly increasing, so KME-CLIP and
KME-no-log rank candidates identically and yield identical retrieval metrics;
they differ only in calibration, that is, in how closely the score matches
$\operatorname{PMI}$ itself. Thus, for rank-based downstream metrics, any difference between the
separately trained variants arises through training; their numerical logits
and induced probabilities nevertheless differ even at fixed parameters.

The two evaluations therefore answer different questions. Downstream accuracy
records how a similarity orders candidates after training in one particular
regime; it is not a measurement of how closely the learned similarity
approximates $\operatorname{PMI}$. The latter is what
\Cref{thm: pmi_error_to_loss,thm: pmi_in_rkhs,thm: discrete_error_2} concern,
and it is what the known-PMI experiments measure directly
(\Cref{app:known-pmi}). There the outer logarithm is the better choice on both
fidelity metrics at $d\in\{8,16\}$, which is the role our analysis assigns it.
The lower-dimensional results at $d\le4$ are mixed. Theorem 7 is an
existence result and does not predict this empirical dimension-wise ordering,
so we report the low-dimensional behavior as an empirical limitation without
offering a mechanism.}
\section{Related Work}
\label{section: related work}
We summarize previous work for improving CLIP's capabilities in this section.
\paragraph{Modification of data.}
To address CLIP's requirement for massive training data, \cite{li2022declip} proposed DeCLIP, which incorporates self-supervised learning techniques, probabilistic generation of multiple captions and images through augmentation, and learning from semantically similar captions.
\cite{gokhale2022spatial} introduced specialized datasets to enhance spatial understanding capabilities, an area where CLIP demonstrates notable limitations.
\cite{parashar2024neglected} developed an algorithm to mitigate class imbalances in web-collected training data.

\paragraph{Modification of loss function.}
\cite{yao2022filip} proposed FILIP, which maximizes similarity at the token level.
\cite{mu2022slip} proposed SLIP, which uses in-modal self-supervised learning in addition to the usual contrastive loss.
\cite{goel2022cyclip} proposed CyCLIP, which incorporates geometric structure into the alignment of image and text embeddings.
\cite{gong2025kernel} proposed an algorithm that uses kernel functions to train image embeddings from different pretrained encoders to become more similar to each other.
\cite{dong2023mask} proposed MaskCLIP, which utilizes distilled representations from masked images.

\paragraph{Modification of modeling.}
Our method falls into this category.
\cite{NEURIPS2022_cloob} proposed CLOOB, which employs a Hopfield network for similarity calculation, based on the analogy that humans extract such relationships through memory.
\cite{desai2023hyperbolic} introduced MERU, which embeds representations in hyperbolic space to better capture hierarchical structures.
\cite{chou2024geometries} proposed EuCLIP, which can capture hierarchical structures without using a geometry of hyperbolic space like MERU.
\cite{uesaka2025wpse} proposed WPSE, which is closely related to our approach;
we have discussed the distinctions between our method and WPSE in \Cref{section: comparison}.

\paragraph{Other related work.}
\cite{ji2023power} provided a theoretical analysis of a linear variant of the contrastive loss in finite dimensions.
Finally, while not on contrastive learning,
\cite{nishikawa2025freedom} pointed out that
softmax operations in the attention mechanism can be treated within the framework of positive definite kernels, similar to our \Cref{prop: kme_clip-clip}.

% \paragraph{Theoretical analysis of contrastive learning}
% \cite{ji2023power} analyzed the properties of linear representation learning. 
% {\color{red}[Isn't WPSE going beyond modification of loss functions?? Maybe we can classify it as modification of architecture??]}

\section{Conclusion}
We have proposed a novel approach for refining similarity computation in CLIP, termed KME-CLIP, which leverages kernel mean embeddings in a reproducing kernel Hilbert space. 
Our theoretical analysis demonstrates that
\blue{the pre-log RKHS score in KME-CLIP can approximate
$\exp(\mathrm{PMI})$ with arbitrary accuracy under the stated assumptions}
as the size of point set increases.
\blue{From an empirical perspective, we evaluate KME-CLIP against CLIP and
WPSE in retrieval and zero-shot classification tasks, and additionally report
a matched KME-no-log control that isolates the outer logarithm
(\Cref{sec: outer-log}). The two evaluations answer different questions: the
control is stronger on CC3M downstream metrics, whereas the logarithm gives
the more faithful approximation of PMI where the ground truth is available.}

{
As future work, we highlight several important directions. First, it would be valuable to explore the use of pretrained backbones for initializing the encoders and to evaluate our method in such settings. In this paper, we intentionally use non-pretrained transformers in all experiments to ensure fair comparison with prior methods. Second, extending our approach to modality pairs beyond image–text is an important direction for assessing its generality.}

\blue{
\subsubsection*{Broader Impact Statement}
Our models are trained on web-scraped image--text data (CC3M and CC12M) and inherit the representational and societal biases present in such corpora.
This concern is not specific to KME-CLIP, which modifies the similarity function rather than the training data, but it applies to any downstream deployment of the resulting representations.
}

\subsubsection*{Acknowledgments}
We are grateful to WeiHsiang Liao, Joan Serrà, and the anonymous reviewers for their constructive feedback, which significantly improved the paper.

\bibliography{main}
\bibliographystyle{tmlr}

% \newpage
\appendix
\onecolumn
\section{Additional information for experiments}\label{appendix: additional_info_ex}
\subsection{Experimental setup}\label{appendix: ex_additional}
\paragraph{Implementation of KME-CLIP.}
To generate multiple point set image embeddings $\{f_i^\mX(x)\}_{i=1}^{m_{\mX}}$, we linearly project the $197 \times 768$ intermediate features of a non-pretrained ViT-B/16 encoder \citep{dosovitskiy2020image} into a $197 \times 512$ tensor and apply $L^2$-normalization to each embedding.
Consequently, we set $m_\mX$ to $197$ and $d$ to $512$.
To derive the positive weights $\{w_i^\mX(x)\}_{i=1}^{m_{\mX}}$ for image embeddings, we linearly project the $197 \times 768$ intermediate features of ViT into a $197 \times 1$ tensor, \blue{pass it through the bounded activation $100\tanh(\cdot/100)$ for numerical stability,} and \blue{then through} a softplus function, \textit{i.e.}, $\log(1+e^{x})$, to ensure positivity.
For the text embeddings $\{f_i^\mY(y)\}_{i=1}^{m_{\mY}}$, we project the $77 \times 512$ output tensor from a non-pretrained Transformer encoder \citep{vaswani2017attention} with width $512$, $12$ layers, and $8$ attention heads into a $77 \times 512$ tensor via linear projection and apply $L^2$-normalization to each embedding.
Accordingly, we set $m_\mY$ to $77$.
Regarding the text positive weights $\{w_i^\mY(y)\}_{i=1}^{m_{\mY}}$, we linearly project the $77 \times 512$ output tensor of the Transformer into a $77 \times 1$ tensor and pass it through \blue{the same two activations}.
We employ the Gaussian kernel for the non-linear kernel function $k$, \textit{i.e.}, $k(u,v):=\exp(-\nor{u-v}^2/2\sigma^2)$. To avoid numerical instability when $\sigma$ approaches zero, we \blue{learn an unconstrained scalar $b=1/\sigma$ and use $b^2=1/\sigma^2$ in the Gaussian kernel}.
\blue{KME-CLIP has no temperature parameter of its own; the inverse length $b$ is its only scale parameter.}

\paragraph{Implementation of CLIP.}
For the image embedding of CLIP, we projected the $768$ dimensional output vector of ViT (it is a projected patch at the position of [CLS] token) into $512$ dimensional tensor by a linear projection.
For the text embedding of CLIP, we only used the [EOS] embedding of the Transformer and projected it into the $512$ dimensional vector by a linear projection.
\blue{The temperature $\tau$ of the CLIP loss is trained jointly with the encoders, initialized at $1/\tau\simeq14.3$ (that is, $\tau=0.07$) following common practice, and constrained to $1/\tau\in[1,100]$ during training.}

\paragraph{Implementation of WPSE.}
As for WPSE, we followed the setting in \cite{uesaka2025wpse}.
Moreover, the procedure for obtaining the image and text embeddings $\{f_i^\mX(x)\}_{i=1}^{m_{\mX}}$, $\{f_i^\mY(y)\}_{i=1}^{m_{\mY}}$ and weights $\{w_i^\mX(x)\}_{i=1}^{m_{\mX}}$, $\{w_i^\mY(y)\}_{i=1}^{m_{\mY}}$ is identical to KME-CLIP except for passing through a softplus function\blue{, so that the WPSE weights are not constrained to be positive}.
\blue{WPSE keeps the loss form of CLIP and replaces the inner product of the two
Euclidean embeddings by the RKHS inner product of the two weighted point sets.
Its temperature $\tau$ is trainable and initialized at
$1/\tau\simeq14.3$, as in our CLIP baseline; unlike the CLIP baseline,
however, it is not explicitly clamped during WPSE training in our implementation.}
For the kernel function $k$, we combined Gaussian kernel and linearized kernel using linear combination\blue{, with the coefficients $\frac13$ and $\frac23$ of \eqref{eq:wpse-kernel}}.
To reduce computational cost, we utilized the random Fourier feature\blue{, with $1024$ features,} for calculating the kernel function.
\blue{The bandwidth of the Gaussian component is fixed at $\sigma_{\mathrm{WPSE}}=0.5$ and is not learned.
The random Fourier
frequencies are resampled for each minibatch from the distribution determined
by that bandwidth.
WPSE therefore has a trained temperature and a fixed kernel bandwidth, whereas KME-CLIP has a trained bandwidth and no temperature.}

\paragraph{Details of hyper-parameter.}
For pretraining, we used the following hyper-parameter presented in Table \ref{table: hyper-param} across CLIP, WPSE, and KME-CLIP on both CC3M and CC12M. For the initialization of the \blue{inverse-length parameter $b$} in
the Gaussian kernel, we set the initial value of $1/\sigma$ in KME-CLIP to $3.78$\blue{, so that $b^2=1/\sigma^2\simeq14.3$. Through the correspondence $\sigma^2=\tau$ of \Cref{prop: kme_clip-clip}, this matches the initialization $1/\tau\simeq14.3$ used for CLIP and WPSE, so all three methods start from the same scale.}

\begin{table*}[htbp]
    \caption{Hyper-parameter for pre-training the encoders}
    \label{table: hyper-param}
    \begin{center}
    \begin{tabular}{lr} \toprule
       Config & Value \\ \midrule
       Total epochs & 50 \\
       Warmup epochs & 2 \\ 
       Warmup start learning rate & $10^{-6}$ \\
       Warmup end learning rate & $5.0\times10^{-4}$ \\
       End learning rate & $10^{-5}$ \\       
       AdamW $\beta$ & $\beta_1,\beta_2=(0.9, 0.98)$ \\       
       AdamW $\epsilon$ & $10^{-8}$ \\
       AdamW weight decay & $0.5$ \\ \bottomrule
    \end{tabular}
    \end{center}
    \vskip -0.2in
\end{table*}

\subsection{Linear Classification}\label{section: linear_classification}
Linear classification is the image classification by logistic regression using features generated by the trained encoder.
We evaluated the image encoder's linear classification ability by the same datasets as the zero-shot classification.
For CLIP, we use the $768$ dimensional output vector of ViT as the feature vector for the logistic regression.
For KME-CLIP and WPSE, we extracted the $197\times 768$ intermediate features from ViT and calculated the $768$ dimensional weighted vector, using the learned weight. Then we use this feature vector as the logistic regression. This procedure is also done in SLIP \citep{mu2022slip}.

Following \cite{uesaka2025wpse}, we ran hyperparameter searches over $C \in [10^{-6}, 10^6]$ with a parametric binary search on a validation split of each dataset. We trained a classifier on the combination of training and validation splits and report its performance on the test split.

The result is shown in Table \ref{table: linear}.
\blue{KME-CLIP is comparable to the baselines under CC3M pretraining, but it is the weakest of the three under CC12M pretraining ($78.31$ versus $81.17$ for WPSE and $79.45$ for CLIP), so we do not claim that it yields better linearly separable features. One relevant difference is that the probe uses features extracted before projection into the RKHS, so it does not evaluate the similarity function that our method modifies; we note this rather than assert it as the explanation of the gap.}

\begin{table}[htbp]
\caption{Linear classification results. Mean per-class accuracy (\%) is reported for Caltech-101, Aircraft, Flowers, and Pets datasets. Top-1 accuracy (\%) is used for all other datasets.
%Models were trained on CC3M and CC12M.
%Gray highlighting indicates the models based on weighted point clouds.
}
\label{table: linear}
%\vskip 0.15in
\begin{center}
\begin{scriptsize}
\tabcolsep = 3pt
\begin{tabular}{cccccccccccccccc}
\toprule
%\multicolumn{2}{c}{Model}
& Model & \rotatebox{90}{Average} &
\rotatebox{90}{ImageNet} & \rotatebox{90}{\tiny CIFAR-10} & \rotatebox{90}{\tiny CIFAR-100} & \rotatebox{90}{STL-10} & \rotatebox{90}{Food-101} & \rotatebox{90}{\tiny Caltech-101} & \rotatebox{90}{Cars} & \rotatebox{90}{Aircraft} &
\rotatebox{90}{Flowers} & \rotatebox{90}{EuroSAT} &
\rotatebox{90}{DTD} & \rotatebox{90}{Pets} &
\rotatebox{90}{SUN397} \\
\midrule
\multirow{3}{*}{CC3M} 
& CLIP  & 72.29 & 59.59 & 88.75 & 70.33 & 92.94 & 67.02 & 82.34 & 37.45 & 44.20 & 92.02 & \textbf{95.88} & 66.01 & 72.88 & 70.32 \\
& WPSE & 74.29 & 61.34 & 89.51 & 70.89 & \textbf{93.63} & \textbf{69.80} & \textbf{85.22} & 44.35 & 48.51 & 92.57 & 95.10 & \textbf{68.03} & \textbf{75.22} & \textbf{71.64} \\
&  KME-CLIP  & \textbf{74.44} & \textbf{62.02} & \textbf{89.73} & \textbf{70.94} & 93.10 & 69.76 & 84.62 & \textbf{45.69} & \textbf{49.28} & \textbf{93.16} & 95.80 & 67.82 & 74.64 & 71.13 \\
\midrule
\multirow{3}{*}{CC12M}
& CLIP & 79.45 & 67.79 & 91.23 & 73.96 & 95.56 & 78.79 & 88.50 & 66.55 & 48.77 & 93.46 & 95.74 & \textbf{74.41} & 81.58 & 76.50  \\
& WPSE & \textbf{81.17} & \textbf{69.71} & \textbf{91.98} & 74.78 & \textbf{95.69} & \textbf{79.81} & \textbf{89.93} & \textbf{71.45} & \textbf{57.60} & \textbf{94.93} & 96.02 & 72.66 & \textbf{83.62} & \textbf{77.05} \\
&  KME-CLIP & 78.31 & 67.76 & 91.92 & \textbf{75.24} & 95.41 & 78.28 & 88.13 & 60.22 & 46.54 & 93.52 & \textbf{96.16} & 73.72 & 74.89 & 76.20  \\
\bottomrule
\end{tabular}
\end{scriptsize}
\end{center}
%\vskip -0.1in
\vspace{-10pt}
\end{table}

% \begin{table*}[htbp]
%     \begin{center}
%     \caption{Hyper-parameter for pre-training the encoders}
%     \label{table: hyper-param}
%     \begin{tabular}{lrr} \hline
%        Config & CC3M & CC12M \\ \hline
%        Epochs & 50 & 50 \\
%        Warmup epochs & 2 & 2 \\ 
%        Warmup start lr & $10^{-6}$ & $10^{-6}$ \\
%        Warmup end lr & $5.0\times10^{-4}$ & $5.0\times10^{-4}$ \\
%        End lr & $10^{-5}$ & $10^{-5}$ \\       
%        AdamW betas & $(0.9, 0.98)$ & $(0.9, 0.98)$ \\       
%        AdamW eps & $10^{-8}$ & $10^{-8}$ \\
%        AdamW weight decay & $0.5$ & $0.5$ \\ \hline
%     \end{tabular}
%     \end{center}
%     \vskip -0.2in
% \end{table*}

\subsection{Computational cost}\label{section: computational_cost}
To address computational efficiency concerns, we also investigate the performance of our method when reducing the size of point set $m_\mX$ of the image encoder.

For training costs, we measured the number of H100 GPUs required and the time per epoch when training on CC3M. 
For inference costs, we measured the time required to complete the retrieval task on the validation split of CC3M using two H100 GPUs.
We also report the top-1 accuracy for each configuration.
The experimental settings follow those described in \Cref{section: ablation}.

The inference-related results for trained models are presented in Table \ref{table: inference_cost},
and the training resource and cost for each configuration is summarized in Table \ref{table: training_cost}.
% Notably, KME-CLIP generally outperforms CLIP even when operating with a reduced size of point set.
% When we set the size of point set to $10$, KME-CLIP maintains superior performance compared to CLIP while requiring the same number of GPUs and nearly identical training time.
% \blue{In this single run, KME-CLIP is above the CLIP baseline at every reduced
% point-set size, although at $m_\mX=2$ the margin ($23.71$ versus $23.66$) is
% $0.05$ points and is smaller than the seed-to-seed variation we observe in
% \Cref{app:real-data-nolog}; we do not read it as a meaningful gap. At
% $m_\mX=10$, KME-CLIP is above CLIP by $0.27$ points while using the same number
% of GPUs and nearly the same per-epoch time.}
In this single run, KME-CLIP is above the CLIP baseline at every reduced
point-set size, although at $m_\mX=2$ the margin ($23.71$ versus $23.66$) is
$0.05$ points and is smaller than the seed-to-seed variation we observe in
\Cref{app:real-data-nolog}; we do not read it as a meaningful gap. At
$m_\mX=10$, KME-CLIP is above CLIP by $0.27$ points while using the same number
of GPUs and nearly the same per-epoch time. The same caveat applies to this
margin: it is also smaller than the seed-to-seed variation in
\Cref{app:real-data-nolog}, so we use the $m_\mX=10$ configuration as a
like-for-like cost anchor rather than as evidence of an accuracy gain.

\blue{Table~\ref{table: training_cost} is not a fixed-hardware comparison. The
number of H100 GPUs varies across configurations (from two to eight) because
larger point sets require more memory, so the reported per-epoch wall-clock
times are not directly comparable across columns: the apparent speed-up at
$m_\mX=100$ and $m_\mX=197$ reflects the larger GPU count rather than a lower
computational cost. A hardware-normalized comparison would require rerunning
every configuration on a fixed number of devices, which we did not perform. The
configurations that share the same hardware are directly comparable, and the
$m_\mX=10$ comparison noted above is one of these.}

\begin{table*}[htbp]
    \caption{Top1 accuracy (\%) and inference time required, averaged over text and image retrieval on validation split of CC3M.}
    \label{table: inference_cost}
    \begin{center}
    \begin{tabular}{lrrrrrr}
        \toprule
       Image point-set size $m_\mX$ & CLIP & 2 & 10 & 50 & 100 & 197 \\
       \midrule
        Accuracy (\%) & 23.66 & 23.71 & 23.93 & 24.35 & 24.79 & 25.57 \\ 
        Inference time with two GPUs (sec) & 126 & 138 & 140 & 159 & 186 & 257  \\ \bottomrule
    \end{tabular}
    \end{center}
    \vskip -0.2in
\end{table*}

\begin{table*}[htbp]
    \caption{Number of H100 GPUs and time required for training by KME-CLIP and CLIP on CC3M.}
    \label{table: training_cost}
    \begin{center}
    \begin{tabular}{lrrrrrr}
        \toprule
       Image point-set size $m_\mX$ & CLIP & 2 & 10 & 50 & 100 & 197 \\
       \midrule
       Training time (sec/epoch) & 1604 & 1413 & 1439 & 925 & 722 & 945 \\
       Number of GPUs for training & 2 & 2 & 2 & 4 & 8 & 8 \\ \bottomrule
    \end{tabular}
    \end{center}
    \vskip -0.2in
\end{table*}

\subsection{Details of the LogSim diagnostic}
\label{app: logsim}

We define the LogSim diagnostic used in Section~\ref{sec: experiment-gap}.
For $A,B\in\{\mX,\mY\}$, let
\[
    K_\theta^{AB}(a,b)
    :=
    \ip{h_\theta^A(a),h_\theta^B(b)}_\mH
\]
denote the pre-log RKHS inner product. Thus, $K_\theta^{\mX\mY}$ is the cross-modal RKHS inner product, while $K_\theta^{\mX\mX}$ and $K_\theta^{\mY\mY}$ are the corresponding in-modal RKHS inner products.

For the Gaussian kernel
\[
    k_\sigma(u,v)
    =
    \exp\!\left(
        -\frac{1}{2\blue{\sigma^2}}\|u-v\|^2
    \right),
\]
we use the constant-corrected log-similarity
\[
    \bar S_\theta^{AB}(a,b)
    :=
    \log K_\theta^{AB}(a,b)+\blue{\frac{1}{\sigma^2}}.
\]
The additive correction $+\blue{1/\sigma^2}$ corresponds to the constant appearing in Proposition~\ref{prop: kme_clip-clip}. It cancels in the contrastive loss, but it must be restored for cosine-style diagnostic normalization.

We then define
\[
    \mathrm{LogSim}_\theta(x,y)
    :=
    \frac{
        \bar S_\theta^{\mX\mY}(x,y)
    }{
        \sqrt{
            \bar S_\theta^{\mX\mX}(x,x)
            \bar S_\theta^{\mY\mY}(y,y)
        }
    }.
\]
In Table~\ref{tab:modality_gap}, we report $\E[\mathrm{LogSim}_\theta(X,Y)]$ over true image--text pairs.

This diagnostic is consistent with the one-point CLIP special case.
Indeed, when $m_\mX=m_\mY=1$ and $w_1^\mX=w_1^\mY=1$, the proof of Proposition~\ref{prop: kme_clip-clip} gives
\[
    \log K_\theta^{\mX\mY}(x,y)
    =
    -\blue{\frac{1}{\sigma^2}}
    +
    \blue{\frac{1}{\sigma^2}} f_1^\mX(x)^\top f_1^\mY(y)
\]
for $L^2$-normalized embeddings. Therefore, we have
\[
    \bar S_\theta^{\mX\mY}(x,y)
    =
    \blue{\frac{1}{\sigma^2}} f_1^\mX(x)^\top f_1^\mY(y),\qquad
    \bar S_\theta^{\mX\mX}(x,x)
    =
    \bar S_\theta^{\mY\mY}(y,y)
    =
    \blue{\frac{1}{\sigma^2}}.
\]
Hence
\[
    \mathrm{LogSim}_\theta(x,y)
    =
    f_1^\mX(x)^\top f_1^\mY(y),
\]
which is exactly the CLIP cosine score.
\section{Proof of auxiliary results}\label{section: auxiliary}

\subsection{Proof of Proposition \ref{prop: kme_clip-clip}}
\begin{proof}
Under the given setting, the similarity of KME-CLIP is
\begin{align}
    S(x,y)=&\log\sum_{i,j} w_i^{\mX}(x)w_j^{\mY}(y) k(f_i^{\mX}(x), f_j^{\mY}(y)) \\
    =&\log (\exp(-\nor{f_1^{\mX}(x)-f_1^{\mY}(y)}^2 / 2\sigma^2)) \\
    =&\frac{-\nor{f_1^{\mX}(x)}^2-\nor{f_1^{\mY}(y)}^2+2f_1^{\mX}(x)^\top f_1^{\mY}(y)}{2\sigma^2} \\
    =&-\frac{1}{\sigma^2}+f_1^{\mX}(x)^\top f_1^{\mY}(y)/\sigma^2,
\end{align}
where the last equality derives from the $L^2$ normalization when calculating the embedding. 
Since we set $\tau=\sigma^2$, taking $g^\mX=f_1^\mX$ and $g^\mY=f_1^\mY$ concludes the desired.
\end{proof}
\section{Proof of theorems in Section \ref{section: theory}}\label{section: proof}

\subsection{Proof of Theorem~\ref{thm: pmi_error_to_loss}}\label{section: proof-thm3}
We start from the following lemma.
\begin{lemma}\label{lem: error-estimate}
    Let $\gS$ be a measure space associated with a finite measure $\mu$.
    Let $\epsilon\ge \delta>0$ and $f,g:\gS\to\R_+$ satisfy the following:
    \begin{itemize}
        \item $\epsilon \le 1/e^2$,
        \item $f$ is a density function with respect to $\mu$,
        \item $g$ satisfies $\lvert f(s)-g(s) \rvert \le \epsilon$
            and $g(s)\ge \delta$ for all $s\in\gS$.
    \end{itemize}
    Then, we have
    \[
        \left\lvert\int_\gS f(s) (\log f(s) - \log g(s)) \dd\mu(s)\right\rvert
        \le 2\sqrt{\epsilon} + \mu(\gS)\sqrt{\epsilon}\log\left(\frac1{\epsilon} + \frac1{\delta}\right).
    \]
\end{lemma}
\begin{proof}
    Let $A=\{s\in\gS\mid f(s)\le\sqrt\epsilon \}$.
    We decompose the integral into three terms:
    \begin{align}
        &\int_\gS f(s) (\log f(s) - \log g(s)) \dd\mu(s)\nonumber\\
        &= \int_A f(s)\log f(s)\dd\mu(s)
        - \int_A f(s)\log g(s)\dd\mu(s)
        + \int_{\gS\setminus A} f(s)(\log f(s) - \log g(s))\dd\mu(s).
        \label{eq: logfg-decomp}
    \end{align}
    For the first term, since $\sqrt\epsilon < 1/e$ and $\lvert x\log x\rvert$ is monotonically increasing over $(0, 1/e)$,
    we have
    \begin{align}
        \left\lvert\int_A f(s)\log f(s)\dd\mu(s)\right\rvert
        \le \int_A \lvert\sqrt{\epsilon}\log\sqrt{\epsilon}\rvert\dd\mu(s)
        \le \frac{\mu(\gS)}2\sqrt{\epsilon}\log\frac1\epsilon.
        \label{eq: first-term-kl}
    \end{align}
    For the second term,
    for each $s\in A$, we have
    $\delta \le g(s) \le f(s) + \epsilon \le \sqrt{\epsilon} + \epsilon \le 1/e + 1/e^2 <1$.
    Thus, we obtain
    \begin{align}
        \left\lvert\int_A f(s)\log g(s)\dd\mu(s)\right\rvert
        \le \int_A f(s)\log\frac1\delta\dd\mu(s)
        \le {\mu(\gS)}\sqrt{\epsilon}\log\frac1\delta.
        \label{eq: second-term-kl}
    \end{align}
    For the final term of \eqref{eq: logfg-decomp}, we need to bound
    $\lvert \log f(s) - \log g(s)\rvert$.
    For a positive constant $a>0$,
    $\log (a+x)$ is $(1/a)$-Lipschitz continuous for $x\ge0$.
    By using this fact, since $f(s)\ge\sqrt{\epsilon}$ and $g(s)\ge \sqrt{\epsilon}-\epsilon$
    for $s\in\gS\setminus A$,
    we have
    \[
        \lvert \log f(s) - \log g(s)\rvert \le \frac{\lvert f(s) - g(s)\rvert}{\sqrt{\epsilon}-\epsilon}
        \le \frac{\epsilon}{\sqrt{\epsilon}-\epsilon}
        = \frac{\sqrt{\epsilon}}{1 - \sqrt{\epsilon}}
        \le \frac{\sqrt{\epsilon}}{1 - 1/e} < 2\sqrt{\epsilon}.
    \]
    Therefore, we have
    \begin{align}
        \left\lvert\int_{\gS\setminus A} f(s)(\log f(s) - \log g(s))\dd\mu(s)\right\rvert
        \le \int_{\gS\setminus A} f(s)2\sqrt{\epsilon}\dd\mu(s)
        \le 2\sqrt{\epsilon}.
        \label{eq: third-term-kl}
    \end{align}

    By applying \eqref{eq: first-term-kl}--\eqref{eq: third-term-kl} to \eqref{eq: logfg-decomp}
    and rearranging some constants,
    we obtain the desired estimate.
\end{proof}

Let us now prove the theorem.
\begin{proof}[Proof of \Cref{thm: pmi_error_to_loss}]
    From \Cref{thm: pmi-minimzer},
    $S = \log\frac{p(x,y)}{p(x)p(y)}$ attains the minimum of $L_S$.
    Let $f(x, y)=\frac{p(x,y)}{p(x)p(y)}$.
    We \blue{first} estimate the error for first term of $L_S$:
    \begin{align}
        &\left\lvert\mathbb{E}_{p(x,y)}\!\left[
                \log \frac{f(x,y)
            }{
                \mathbb{E}_{p(x')}[f(x',y)]
            }\right]
        - \mathbb{E}_{p(x,y)}\!\left[
                \log \frac{h(x,y)
            }{
                \mathbb{E}_{p(x')}[h(x',y)]
        }\right]\right\rvert
        \nonumber\\
        &\le \left\lvert\mathbb{E}_{p(x,y)}[\log f(x, y) - \log h(x, y)]\right\rvert
            + \left\lvert\mathbb{E}_{p(y)}[\log(\mathbb{E}_{p(x)}[f(x, y)])
            - \log(\mathbb{E}_{p(x)}[h(x, y)])]\right\rvert.
        \label{eq: L_S decomp}
    \end{align}
    By rewriting the first term of \eqref{eq: L_S decomp} by integral,
    we obtain the following:
    \begin{align}
        &\left\lvert\mathbb{E}_{p(x,y)}[\log f(x, y) - \log h(x, y)]\right\rvert\nonumber\\
        &=\left\lvert\iint_{\mX\times\mY}p(x, y)\log f(x, y)\dd\nu_X(x)\dd\nu_Y(y)
        - \iint_{\mX\times\mY}p(x, y)\log h(x, y)\dd\nu_X(x)\dd\nu_Y(y)\right\rvert\nonumber\\
        &=\left\lvert\iint_{\mX\times\mY}f(x, y)\log f(x, y)\dd\tilde{\nu}_X(x)\dd\tilde{\nu}_Y(y)
        - \iint_{\mX\times\mY}f(x, y)\log h(x, y)\dd\tilde{\nu}_X(x)\dd\tilde{\nu}_Y(y)\right\rvert,
        \nonumber
    \end{align}
    where $\mathrm{d}\tilde{\nu}_X(x) = p(x)\dd\nu_X(x)$ and $\mathrm{d}\tilde{\nu}_Y(y) = p(y)\dd\nu_{\blue{Y}}(y)$.
    \blue{Note} also that we have
    \[
        \tilde{\nu}_X(\mX) = \int_\mX p(x)\dd\nu_X(x)=1,
        \qquad
        \tilde{\nu}_Y(\mY) = \int_{\blue{\mY}} p(y)\dd\nu_Y(y)=1,
    \]
    since $p(x)$ and $p(y)$ are density functions,
    and $f(x, y)$ is a density function with respect to $\tilde{\nu}_X\otimes\tilde{\nu}_Y$, because
    \[
        \iint_{\mX\times\mY}f(x, y)\dd\tilde{\nu}_X(x)\dd\tilde{\nu}_Y(y)
        =\iint_{\mX\times\mY}p(x, y)\dd{\nu}_X(x)\dd{\nu}_Y(y)=1.
    \]
    Thus, by applying \Cref{lem: error-estimate} for the product measure $\tilde{\nu}_X\otimes\tilde{\nu}_Y$, we have
    \begin{align}
        \left\lvert\mathbb{E}_{p(x,y)}[\log f(x, y) - \log h(x, y)]\right\rvert
        &\le 2\sqrt{\epsilon} + \tilde{\nu}_X(\mX)\tilde{\nu}_Y(\mY)\sqrt{\epsilon}\log\left(\frac1{\epsilon} + \frac1{\delta}\right)\nonumber\\
        &= 2\sqrt{\epsilon} + \sqrt{\epsilon}\log\left(\frac1{\epsilon} + \frac1{\delta}\right).
        \label{eq: kl-thm-first-term}
    \end{align}
    
    Next, let us estimate the second term of \eqref{eq: L_S decomp}.
    Let $F_Y(y):=\mathbb{E}_{p(x)}[f(x, y)]$ and $H_Y(y):=\mathbb{E}_{p(x)}[h(x, y)]$.
    Since they are expectations, we can see that the assumed inequality constraints is inherited to $H_Y$ and $F_Y$,
    \textit{i.e.}, $\lvert H_Y(y) - F_Y(y)\rvert \le \epsilon$.
    Note also that $F_Y(y)=1$ holds almost everywhere with respect to $\tilde{\nu}_Y$.
    Indeed, for any measurable set $A\subset\mY$, we have
    \begin{align*}
        \int_A F_Y(y)\dd\tilde{\nu}_Y(y)
        &= \int_A p(y)\int_\mX p(x) f(x, y)\dd\nu_X(x)\dd\nu_Y(y) \\
        &= \iint_{\mX\times A} p(x, y)\dd(\nu_X\otimes\nu_Y)(x, y).\\
        &= \mathbb{P}(Y \in A) = \int_A p(y)\dd\nu_Y(y).
    \end{align*}
    Thus, $H_Y(y)\in [1-\epsilon, 1+\epsilon]$ almost everywhere with respect to $\tilde{\nu}_{\blue{Y}}$.
    In particular, we have an integral estimate $\int_\mY\lvert \log H_Y(y)\rvert\dd\tilde{\blue{\nu}}_Y(y)
    \le \log\frac1{1-\epsilon}$.
    By using these, we can evaluate the second term of \eqref{eq: L_S decomp} as follows:
    \begin{align*}
        &\left\lvert\mathbb{E}_{p(y)}[\log(\mathbb{E}_{p(x)}[f(x, y)])
            - \log(\mathbb{E}_{p(x)}[h(x, y)])]\right\rvert\\
        &=\left\lvert\mathbb{E}_{p(y)}[\log(F_Y(y))
            - \log(H_Y(y))]\right\rvert\\
        &=\left\lvert\int_{\mY}p(y)\log F_Y(y)\dd{\nu}_Y(y)
        - \int_{\mY}p(y)\log H_Y(y)\dd{\nu}_Y(y)\right\rvert\\
        &=\left\lvert\int_{\mY}\log H_Y(y)\dd\tilde{\nu}_Y(y)\right\rvert\\
        &\le \int_\mY\lvert \log H_Y(y)\rvert\dd\tilde{\blue{\nu}}_Y(y)
        \le \log\frac1{1-\epsilon}.
    \end{align*}
    Since $\frac{\mathrm{d}}{\mathrm{d}x}\log\frac1{1-x}=\frac1{1-x}$, the function
    $\log\frac1{1-x}$ is $\frac1{1-1/e^2}$-Lipschitz continuous for $0\le x\le 1/e^2$.
    Thus, we have
    \begin{align}
        &\left\lvert\mathbb{E}_{p(y)}[\log(\mathbb{E}_{p(x)}[f(x, y)])
            - \log(\mathbb{E}_{p(x)}[h(x, y)])]\right\rvert\nonumber\\
        &\le \log\frac1{1-\epsilon}
        \le \frac1{1-1/e^2}\epsilon
        = \frac{\sqrt{\epsilon}}{1-1/e^2}\sqrt{\epsilon}
        \le \frac{1/e}{1-1/e^2}\sqrt{\epsilon}
        = \frac{1}{e-1/e}\sqrt{\epsilon} \le \sqrt{\epsilon},
        \label{eq: kl-thm-second-term}
    \end{align}
    where we have used the assumption $\epsilon\le1/e^2$ multiple times.
    By applying \eqref{eq: kl-thm-first-term} and \eqref{eq: kl-thm-second-term} to \eqref{eq: L_S decomp},
    we can bound the error of the first term (without $1/2$) of $L_S$ by the right-hand side of \eqref{eq: pmi_erro_to_loss}.
    By conducting the same argument on the second term and taking average,
    we obtain the desired estimate for $L_S$.
\end{proof}

\subsection{Proof of Theorem \ref{thm: pmi_in_rkhs}}\label{section: proof-thm4}
\begin{proof}
    First, since $\int_\gZ \blue{\frac{p(x|z)}{p(x)}}\dd\rho(z)=1$,
    there exists a $z$ such that $\blue{p(x|z)/p(x)}\le 1$.
    Thus, from the Lipschitz continuity and the compactness of $\gZ$,
    $p(x|z)/p(x)$ is bounded by a universal constant.
    We can show the same for $\blue{p(y|z)/p(y)}$,
    so there is a universal constant $C^\prime$ such that
    \[
        \frac{p(x|z)}{p(x)}, \frac{p(y|z)}{p(y)} \le C^\prime.
    \]

    Since $\gZ$ is compact,
    for any $q>0$,
    there is a finite covering $\gZ = B_1\cup\cdots\cup B_m$ such that
    each $B_i$ is an open ball of radius $q/2$ in $\gZ$.
    Let $A_i:=B_i\setminus\bigcup_{j<i}B_j$.
    Then, each $A_i$ is Borel measurable and $\gZ = A_1\cup\cdots\cup A_m$ becomes a disjoint union.
    We assume each $A_i$ is nonempty; otherwise we simply omit such $A_i$.
    % We split the latent space $\gZ$ by spheres whose radius is all $q$. We call each sphere $A_i$ ($i=1,\cdots m$).
    Since ${p(x|z)}/{p(x)}$ and ${p(y|z)}/{p(y)}$ are uniformly $C_L$-Lipschitz continuous with respect to $z$, there exists $a_{A_i}^x, a_{A_i}^y \in (0, C^\prime]$ satisfying
    \begin{align}
        \left|\frac{p(x|z)}{p(x)} - \sum_{i=1}^m a_{A_i}^x \mathbbm{1}[z\in A_i]\right|\le C_Lq,
        \qquad
        \left|\frac{p(y|z)}{p(y)} - \sum_{i=1}^m a_{A_i}^y \mathbbm{1}[z\in A_i]\right|\le C_Lq,
        \label{eq: lip-approx}
    \end{align}
    for all $x,y,z$.
    Indeed, for a $C_L$-Lipschitz function $f$
    and $z, z^\prime\in A_i$,
    we have
    \[
        \lvert f(z) - f(z^\prime)\rvert
        \le C_Ld(z, z^\prime) \le C_L\cdot 2\cdot q/2 = C_Lq,
    \]
    where $d$ is the metric function over $\gZ$.
    By taking any $z_i\in A_i$ for each $i$, we thus have
    \[
        \left\lvert f(z) - \sum_{i=1}^m f(z_i) \mathbbm{1}[z\in A_i]\right\rvert \le C_Lq,
    \]
    because $A_i$ is disjoint to each other.
    This justifies \eqref{eq: lip-approx}. 
    
    Hence, we have
    \begin{align}
        &\left|\int_{\gZ} \frac{p(x|z)p(y|z)}{p(x)p(y)}\dd\rho(z) - \sum_{i=1}^m (a_{A_i}^x a_{A_i}^y)\rho(A_i)\right| \notag \\
        \le& \left|\int_{\gZ}\left(C_Lq+\sum_{i=1}^m a_{A_i}^x \mathbbm{1}[z\in A_i]\right)\left(C_Lq+\sum_{i=1}^m a_{A_i}^y \mathbbm{1}[z\in A_i]\right)\dd\rho(z)-\sum_{i=1}^m(a_{A_i}^x a_{A_i}^y)\rho(A_i)\right| \notag\\
        =&\left|(C_Lq)^2 + C_Lq\sum_{i=1}^m(a_{A_i}^x+a_{A_i}^y)\rho(A_i)+\sum_{i=1}^m(a_{A_i}^x a_{A_i}^y)\rho(A_i)-\sum_{i=1}^m(a_{A_i}^x a_{A_i}^y)\rho(A_i)\right| \notag\\
        =&(C_Lq)^2 + C_Lq\sum_{i=1}^m(a_{A_i}^x+a_{A_i}^y)\rho(A_i).\notag\\
        \le& (C_Lq)^2 + 2C_LC^\prime q\sum_{i=1}^m\rho(A_i)
        =(C_Lq)^2 + 2C_LC^\prime q\label{eq: 1}
    \end{align}

    Let us construct $\mu$ and embedding functions now.
    % We take points $z_i$ in each $A_i$ as $\|z_i-z_j\|\ge q$ holds for $i\neq j$.
    % Let $\mu(z)$ be a discrete probability measure on $\{z_1,\cdots,z_m\}$.
    We take $m$ distinct points $u_i$ in $\R^d$.
    Even if we are constrained in the unit sphere ($d\ge2$),
    we can take $u_i$'s so that, for $i\ne j$
    we have $\lVert u_i - u_j\rVert\ge 2/m$ by manipulating one of the coordinates.
    Let $\mu$ be the uniform measure over $\{u_1,\ldots u_m\}$,
    \textit{i.e.}, $\mu=\frac1m\sum_{i=1}^m \delta_{u_i}$.

    If we set $g_X^x(u_i)=\blue{a_{A_i}^x m\sqrt{\rho(A_i)}}$ and $g_Y^y(u_i)=\blue{a_{A_i}^y m\sqrt{\rho(A_i)}}$
    for $i=1,\ldots,m$,
    they can be regarded as $L^2$ functions with respect to the measure $\mu$.
    Then, we have
     \begin{align}
        &\left|\ip{\blue{\int_{\R^d}g_X^x(u)k(u,\cdot)\dd\mu(u), \int_{\R^d}g_Y^y(u)k(u,\cdot)\dd\mu(u)}}_{\gH}-\sum_{i=1}^m(a_{A_i}^x a_{A_i}^y)\rho(A_i)\right| \notag\\
        =&\left|\ip{\frac1m\sum_{i=1}^m {a_{A_i}^x}\blue{m\sqrt{\rho(A_i)}} k(u_i,\cdot),
        \frac1m\sum_{i=1}^m {a_{A_i}^y}\blue{m\sqrt{\rho(A_i)}} k(u_i,\cdot)}_{\gH}-\sum_{i=1}^m(a_{A_i}^x a_{A_i}^y)\rho(A_i)\right| \notag\\
        =&\left|\sum_{i=1}^m(a_{A_i}^x a_{A_i}^y)\rho(A_i)\blue{k(u_i,u_i)} +
        \sum_{i\neq j}(a_{A_i}^x a_{A_j}^y)\sqrt{\rho(A_i)}\sqrt{\rho(A_j)}\blue{k(u_i,u_j)}-\sum_{i=1}^m(a_{A_i}^x a_{A_i}^y)\rho(A_i)\right| \notag\\
        \le&\sum_{i\neq j}(a_{A_i}^x a_{A_j}^y)\sqrt{\rho(A_i)}\sqrt{\rho(A_j)}
        \exp(-(2/m)^2/(2\sigma^2)) \notag\\
        \le&\blue{\:(C^\prime)^2
        \left(\sum_{i=1}^m\sqrt{\rho(A_i)}\right)^2
        \exp(-2/(m\sigma)^2)} \notag\\
        \le&\blue{\:m(C^\prime)^2\exp(-2/(m\sigma)^2)},\label{eq: 2}
    \end{align}
    where we have used the Cauchy--Schwarz in the last inequality.

    Combining \eqref{eq: 1} and \eqref{eq: 2}, we have
    \begin{align}
        &\left|\ip{\blue{\int_{\mathbb R^d} g_\mX^x(u) k(u,\cdot)\dd\mu(u), \int_{\mathbb R^d} g_\mY^y(u) k(u,\cdot)\dd\mu(u)}}_{\gH}-\frac{p(x,y)}{p(x)p(y)}\right| \notag\\
        \le&(C_Lq)^2 + 2C_LC^\prime q
        +\blue{m(C^\prime)^2}
        \exp(-2/(m\sigma)^2).
        \label{eq: desired upper bound}
    \end{align}
    If we take $q<\frac\epsilon{4\max\{C_L, C_LC^\prime\}}$ and then $\sigma$ to be sufficiently small,
    the right-hand side of \eqref{eq: desired upper bound} becomes
    smaller than $\epsilon$, which concludes the proof.

\end{proof}

\subsection{Proof of Theorem \ref{thm: discrete_error_2}}\label{section: proof-thm5}
\begin{proof}
    We can show the result from straight-forward calculation about the norm in RKHS.
    Let $ f=\int_{\blue{\R^d}} k(u, \cdot)g(u)\dd\mu(u)$. Then,
    for any $u_1,\ldots,u_{m_\mX}\in\R^d$, we have
    \begin{align}
        &\nor{
            f
            - \sum_{i=1}^{m_\mX} \frac{1}{m_\mX} k(u_i, \cdot)g(u_i)
        }_{\gH}^2 \notag\\
        =&\nor{f}_{\gH}^2 + \nor{\sum_{i=1}^{m_\mX} \frac{1}{m_\mX} k(u_i, \cdot)g(u_i)}_{\gH}^2 - 2\ip{f, \sum_{i=1}^{m_\mX} \frac{1}{m_\mX} k(u_i, \cdot)g(u_i)}_{\gH} \notag\\
        =&\iint k(u,u')g(u)g(u')\dd\mu(u)\dd\mu(u')+ \frac{1}{{m_\mX}^2}\sum_{i=1}^{m_\mX}\sum_{j=1}^{m_\mX}k(u_i,u_j)g(u_i)g(u_j) - 2\sum_{i=1}^{m_\mX} \frac{1}{m_\mX}f(u_i)g(u_i) \notag\\
        =&\left(\iint k(u,u')g(u)g(u')\dd\mu(u)\dd\mu(u')-\frac{1}{m_\mX}\sum_{i=1}^{m_\mX} \int k(u_i, u)g(u)g(u_i)\dd\mu(u)\right) \nonumber\notag\\
        +&\left(\frac{1}{{m_\mX}^2}\sum_{i=1}^{m_\mX}\sum_{j=1}^{m_\mX}k(u_i,u_j)g(u_i)g(u_j)-\frac{1}{m_\mX}\sum_{i=1}^{m_\mX} \int k(u_i, u)g(u)g(u_i)\dd\mu(u)\right).
    \label{eq: a}\end{align}

    It is enough to show that the expected value of the right-hand side of \eqref{eq: a} has the upper bound $K\nor{g}_{L^2(\mu)^2}/m_\mX$ when we sample $u_1,...,u_{m_\mX}\sim \mu$.
    
    As for the first term in the right-hand side of \eqref{eq: a}, we have
    \begin{equation*}
        \E\left[\iint k(u,u')g(u)g(u')\dd\mu(u)\dd\mu(u')-\frac{1}{m_\mX}\sum_{i=1}^{m_\mX} \int k(u_i, u)g(u)g(u_i)\dd\mu(u)\right] = 0.        
    \end{equation*}
    Also, as for the second term in the right-hand side of \eqref{eq: a}, we have
    \begin{align*}
        &\E\left[\frac{1}{{m_\mX}^2}\sum_{i,j}k(u_i,u_j)g(u_i)g(u_j)-\frac{1}{m_\mX}\sum_{i=1}^{m_\mX} \int k(u_i, u)g(u)g(u_i)\dd\mu(u)\right] \\
        =&\frac{1}{{m_\mX}^2}\E\left[\sum_{i=1}^{m_\mX} k(u_i,u_i)g(u_i)g(u_i)\right] \\
        +&\frac{1}{{m_\mX}^2}\E\left[\sum_{i\neq j} k(u_i,u_j)g(u_i)g(u_j)\right]-\iint_{\gZ}k(u',u)g(u)g(u')\dd\mu(u)\dd\mu(u') \\
        =&\frac{1}{{m_\mX}^2}\E\left[\sum_{i=1}^{m_\mX} k(u_i,u_i)g(u_i)g(u_i)\right]-\frac{1}{m_\mX}\iint  k(u',u)g(u)g(u')\dd\mu(u)\dd\mu(u').
    \end{align*}
    In the last equality, we used $\E[k(u_i,u_j)g(u_i)g(u_j)]=\iint_{\gZ\times\gZ}k(z,z')g(z)g(z')\dd\mu(z)\dd\mu(z')$ for $i\neq j$,
    which is positive from the positive-definiteness of $k$.
    Thus, the final term is bounded by
    \[
        \frac{1}{{m_\mX}^2}\E\left[\sum_{i=1}^{m_\mX} k(u_i,u_i)g(u_i)g(u_i)\right]
        =\frac1{m_\mX} \int k(u, u)g(u)^2\dd\mu(u)
        \le \frac1{m_\mX}K\nor{g}_{L^2}^2,
    \]
    which is the desired bound.
\end{proof}

\subsection{Proof of Theorem \ref{thm: clip_limitation}}\label{section: proof-thm6}
\begin{proof}
    In this proof, let us write $(i, j)$ as $ij$ for notational simplicity.
    We only have to show that if condition 1. does not hold, i.e., if low modality gap condition holds, then condition 2. holds.
    We give a proof by contradiction, with the assumption that, for each $ij$ and $k\ell$, we have
    \begin{equation}
        \left|\alpha\exp\left(S(ij, k\ell)\right)-\exp(\text{PMI}(ij, k\ell))\right|< \frac{N}4.
        \label{eq: contra-assumption}
    \end{equation}
    
    From \Cref{assumption: 2-mixture}, we have
    $p_X(ij) = p_Y(ij) = \frac2{N(N+1)}$ for all $i,j\in\gZ$.
    Thus, we have
    \begin{equation}
        \frac{p_{X|Z}(ij|\ell)}{p_X(ij)} = \frac{N(N+1)}2 \frac{\delta^{i\ell}+\delta^{j\ell}}{N+1}
        = \frac{N}2(\delta^{i\ell}+\delta^{j\ell}).
        \label{eq: thm6-conditional}
    \end{equation}
    The same holds for $p_Y$.
    Thus, for $i\ne j$, we have identities such as
    \begin{equation}
        \exp(\text{PMI}(ii,jj)) = 0,
        \quad \exp(\text{PMI}(ii,ij)) = \frac1N\cdot\frac{2N}2\cdot\frac{N}2=\frac{N}2,
        \quad
        \exp(\text{PMI}(ii,ii)) = N.
        \label{eq: pmi-values}
    \end{equation}
    These also hold for $\text{PMI}^\mX$.
    In particular, given $ii\in\mX$
    we have
    $S(ii, ii) \ge S(ii, j\ell)$ for any $j\ell$
    because of \eqref{eq: contra-assumption}.
    Indeed, $\alpha\exp(S(ii, ii))>3N/4$ should be satisfied from \eqref{eq: contra-assumption} and \eqref{eq: pmi-values},
    but for any other $j\ell$ we have $\alpha\exp(S(ii, j\ell))<3N/4$ from the same reasoning.
    Considering that $\exp(\text{PMI}(ii,jj)) = 0 < \delta$ for any $\delta>0$, we have the following
    from a low modality gap condition:
    \begin{equation}
        S^\mX(ii, jj) \le S(ii, ii).
        \label{eq: similarity-low-modality-gap-inequality}
    \end{equation}
    
    When $u, v\in\R^d$ are on the unit sphere, we have
    \[
        \alpha\exp(u^\top v/\tau) = \alpha\exp\left(\frac{-\lVert u - v\rVert^2 +\nor{u}^2 + \nor{v}^2}{2\tau}\right)
        = \alpha\exp(1/\tau)\exp\left(-\frac{\lVert u - v\rVert^2}{2\tau}\right).
    \]
    Thus, by letting $\alpha_0:=\alpha\exp(1/\tau)$,
    we have
    \begin{align}
        &\alpha\exp(S^\mX(x, x')) = \alpha_0\exp\left(-\frac{\nor{g^\mX(x)- g^\mX(x')}^2}{2\tau}\right), \notag \\
        &\alpha\exp(S(x, y)) = \alpha_0\exp\left(-\frac{\nor{g^\mX(x)- g^\mY(y)}^2}{2\tau}\right).
        \label{eq: similarity-with-distance}
    \end{align}
    From these, we can translate \eqref{eq: similarity-low-modality-gap-inequality} to 
    $\lVert g^\mX(ii) - g^\mX(jj)\rVert \ge \lVert g^\mX(ii) - g^\mY(ii) \rVert$.
    From the triangle inequality, we also have
    \[
        \lVert g^\mX(ii) - g^\mX(jj)\rVert
        \ge \lVert g^\mX(jj) - g^\mY(ii) \rVert - \lVert g^\mX(ii) - g^\mY(ii) \rVert.
    \]
    By taking the average of these two lower bounds, we have
    \begin{align}
        \lVert g^\mX(ii) - g^\mX(jj)\rVert &\ge \frac{\lVert g^\mX(ii) - g^\mY(ii) \rVert + \lVert g^\mX(jj) - g^\mY(ii) \rVert - \lVert g^\mX(ii) - g^\mY(ii) \rVert}2 \nonumber\\
        &=\frac{\lVert g^\mX(jj) - g^\mY(ii) \rVert}2
        = \frac12\sqrt{2\tau\log\frac{\alpha_0}{\alpha\exp(S(jj, ii))}}\nonumber\\
        &\ge \frac12\sqrt{2\tau\log\frac{\alpha_0}{\exp(\text{PMI}(jj, ii)) + N/4}}
        =\frac12\sqrt{2\tau\log\frac{\alpha_0}{N/4}},
        \label{eq: lower-bound-distance}
    \end{align}
    where we have used \eqref{eq: contra-assumption}, \eqref{eq: pmi-values}, and \eqref{eq: similarity-with-distance}.

    We can also bound the opposite direction as
    \begin{align}
        \lVert g^\mX(ii) - g^\mX(jj)\rVert
        &\le \lVert g^\mX(ii) - g^\mY(ij)\rVert + \lVert g^\mX(jj) - g^\mY(ij)\rVert \nonumber\\
        &=\sqrt{2\tau\log\frac{\alpha_0}{\alpha\exp(S(ii, ij))}} + \sqrt{2\tau\log\frac{\alpha_0}{\alpha\exp(S(jj, ij))}} \nonumber\\
        &\le 2\sqrt{2\tau\log\frac{\alpha_0}{\exp(\text{PMI}(ii, ij)) - N/4}}
        = 2\sqrt{2\tau\log\frac{\alpha_0}{N/4}}.\label{eq: upper-bound-distance}
    \end{align}
    Thus, by letting $R:=\sqrt{2\tau\log\frac{\alpha_0}{N/4}}$,
    we have $R/2 \le \lVert g^\mX(ii) - g^\mX(jj)\rVert \le 2R$ from \eqref{eq: lower-bound-distance} and \eqref{eq: upper-bound-distance}.
    Therefore, if we let $B_i$ be an open ball of radius $R/4$ centered at $g^\mX(ii)$,
    $B_1,\ldots, B_N$ are disjoint to each other and included in an ball of radius $(2+\frac14)R$, centered at a certain $g^\mX(\ell\ell)$.
    By summing up the Lebesgue measure of $N$ balls,
    we have $N(1/4)^dS \le (9/4)^dS$, where $S$ is the measure of a unit ball in $\R^d$.
    Therefore, $N$ needs to satisfy $N\le 9^d$.
\end{proof}

\subsection{Proof of Theorem \ref{thm: kme-clip-superiority}}\label{section: proof-thm7}
\begin{proof}
    We follow the notation of the proof of \Cref{thm: clip_limitation}.
    Recall \eqref{eq: thm6-conditional} stating the following:
    \[
        \frac{p_{X|Z}(ij|\ell)}{p_X(ij)} = \frac{p_{Y|Z}(ij|\ell)}{p_Y(ij)}
        = \frac{N}2(\delta^{i\ell}+\delta^{j\ell}).
    \]
    From this, we can write the exponential PMI as follows:
    \begin{align}
        \exp(\text{PMI}(ij, st))
        &= \frac1N\sum_{\ell=1}^N
        \frac{p_{X|Z}(ij|\ell)}{p_X(ij)}
        \blue{\frac{p_{Y|Z}(st|\ell)}{p_Y(st)}}
        \nonumber\\
        &= \frac{N}4\sum_{\ell=1}^N (\delta^{i\ell}+\delta^{j\ell})(\delta^{s\ell}+\delta^{t\ell})
        = \frac{N}4(\delta^{is}+\delta^{it}+\delta^{js}+\delta^{jt}).
        \label{eq: kronecker-expansion}
    \end{align}
    Following the idea of \Cref{section: proof-thm4}, let us take $N$ distinct points
    $u_1,\ldots, u_N$ on the unit sphere of $\R^d$,
    such that $\lVert u_i-u_j\rVert\ge 1/N$ for each $i\ne j$.
    We realize the embedding by
    \[
        h_\theta^\mX(ij) = h_\theta^\mY(ij) = \frac{\sqrt{N}}2(k(u_i, \cdot) + k(u_j, \cdot)).
    \]
    The similarities induced by these RKHS embeddings satisfies a low modality gap condition in Definition \ref{definition: low-modality-gap},
    since $S^\mX(x, x') = S(x, x')$ holds for every $x,x'\in\mX$.
    Then, we have
    \begin{equation}
        \ip{h_\theta^\mX(ij), h_\theta^\mY(st)}_\gH
        = \frac{N}4(k(u_i, u_s) + k(u_i, u_t) + k(u_j, u_s) + k(u_j, u_t)).
        \label{eq: kernel-sum-expansion}
    \end{equation}
    By the term-by-term comparison of \eqref{eq: kronecker-expansion} and \eqref{eq: kernel-sum-expansion},
    proving $\lvert k(u_i, u_\ell) - \delta^{i\ell}\rvert < \epsilon/N$
    for each $i,\ell$ is sufficient for achieving the desired error bound.
    Because $k$ is a Gaussian kernel, we have $k(u_i, u_i)=1$.
    Therefore, it suffices to prove that we can let $k(u_i, u_\ell)<\epsilon/N$ when $i\ne\ell$.
    Since we have
    \[
        k(u_i, u_\ell)=\exp\left(-\frac{\lVert u_i - u_\ell\rVert^2}{2\sigma^2}\right)
        \le \exp\left(-\frac1{2N^2\sigma^2}\right)
    \]
    from the construction,
    just letting $\sigma<(2N^2\log(N/\epsilon))^{-1/2}$ gives the desired conclusion.
\end{proof}

\section{Further ablation and synthetic experiments}

\subsection{Multi-seed ablation with and without logarithmic operation}
\label{app:real-data-nolog}

We additionally compare CLIP, WPSE, KME-no-log, and KME-CLIP after CC3M
pretraining with global batch size $512$ over the same three seeds.
We use a global batch size of $512$ (four devices with a per-device batch
size of $128$) rather than the $1024$ of \Cref{section: experiment}
because running four methods over three seeds at the original batch size was
outside our compute budget. Batch size changes the difficulty of the in-batch
contrastive objective, so the absolute values in this section are not
directly comparable to Tables~\ref{table: retrieval}~and~\ref{table: zero-shot};
we retain the batch-size-$1024$ single-run results in the main paper and report
this matched comparison separately. Within this section the comparison is
controlled: all four methods share the batch size, schedule, data, and seeds.

KME-no-log is matched to KME-CLIP in parameterization---the same encoders and
point-set sizes, the same softplus-parameterized positive weights, and the same
single Gaussian kernel with learned bandwidth and initialization---and changes
only the similarity from
\[
    S_{\mathrm{KME}}(x,y)
    =
    \log K_\theta(x,y)
    \quad\text{to}\quad
    S_{\mathrm{nolog}}(x,y)
    =
    K_\theta(x,y)
    =
    \left\langle
        h_\theta^\mX(x),h_\theta^\mY(y)
    \right\rangle_{\gH}.
\]
Since the logarithm is strictly increasing, the two similarities induce the
same ranking when applied to the same trained model. Here they are trained
separately and reach different parameters, so what this section compares is
the effect of the two training objectives rather than two read-outs of one
model (see \Cref{sec: outer-log}).

KME-no-log has the highest three-seed mean in all $18$ retrieval entries.
Across the $13$ zero-shot datasets, it also has the highest mean accuracy,
$29.18\pm0.17$, compared with $28.68\pm0.96$ for KME-CLIP,
$27.24\pm1.73$ for WPSE, and $26.89\pm0.86$ for CLIP. The dataset-level
ordering is not uniform: KME-no-log has the highest mean on eight datasets,
KME-CLIP on four, and WPSE on one.
These downstream results show that the outer logarithm is not uniformly
beneficial for real-data retrieval and classification. Because downstream
accuracy does not directly measure conditional-distribution or density-ratio
fidelity, \Cref{app:known-pmi} evaluates the same contrast in a setting where
the ground-truth PMI is known.

\begin{table}[htbp]
\centering
\caption{\blue{Matched retrieval results after CC3M pretraining with global batch
size $512$. Values are percentages reported as mean $\pm$ sample standard
deviation over three seeds. Bold indicates the highest mean and underlining
indicates the second-highest mean in each column. Text retrieval denotes
image-to-text retrieval, and image retrieval denotes text-to-image retrieval.}}
\label{table:ablation-retrieval}
\resizebox{\linewidth}{!}{%
\begin{tabular}{@{}lccccccccc@{}}
\toprule
& \multicolumn{3}{c}{CC3M}
& \multicolumn{3}{c}{MSCOCO}
& \multicolumn{3}{c}{Flickr30K} \\
\cmidrule(lr){2-4}\cmidrule(lr){5-7}\cmidrule(lr){8-10}
Model & Top1 & Top5 & Top10 & Top1 & Top5 & Top10 & Top1 & Top5 & Top10 \\
\midrule
Text retrieval \\
CLIP
& $19.17\pm0.75$ & $38.81\pm0.51$ & $48.28\pm0.37$
& $13.16\pm0.46$ & $\underline{32.02\pm0.71}$ & $\underline{43.10\pm0.86}$
& $25.35\pm0.51$ & $51.62\pm0.36$ & $63.31\pm0.10$ \\
WPSE
& $18.60\pm0.87$ & $38.31\pm1.15$ & $47.63\pm1.38$
& $\underline{13.26\pm1.03}$ & $31.37\pm1.71$ & $42.22\pm1.86$
& $24.48\pm2.29$ & $50.64\pm2.14$ & $61.75\pm1.93$ \\
KME-CLIP
& $\underline{20.60\pm0.83}$ & $\underline{40.90\pm1.02}$ & $\underline{50.55\pm1.33}$
& $12.60\pm0.16$ & $31.44\pm0.21$ & $42.84\pm0.34$
& $\underline{26.16\pm0.68}$ & $\underline{53.09\pm0.68}$ & $\underline{64.02\pm0.09}$ \\
KME-no-log
& $\mathbf{20.86\pm0.22}$ & $\mathbf{41.23\pm0.92}$ & $\mathbf{51.02\pm0.83}$
& $\mathbf{16.02\pm0.65}$ & $\mathbf{36.03\pm0.74}$ & $\mathbf{47.17\pm0.92}$
& $\mathbf{30.32\pm0.57}$ & $\mathbf{56.49\pm0.76}$ & $\mathbf{67.67\pm0.63}$ \\
\midrule
Image retrieval \\
CLIP
& $22.61\pm0.36$ & $43.64\pm0.14$ & $53.07\pm0.12$
& $\underline{14.11\pm0.27}$ & $\underline{33.60\pm0.31}$ & $44.62\pm0.33$
& $27.07\pm0.93$ & $52.95\pm0.49$ & $64.29\pm0.29$ \\
WPSE
& $21.80\pm1.11$ & $43.26\pm1.26$ & $52.68\pm1.47$
& $13.93\pm0.91$ & $32.83\pm1.72$ & $43.92\pm1.84$
& $25.54\pm1.55$ & $51.51\pm2.03$ & $64.64\pm3.37$ \\
KME-CLIP
& $\underline{23.29\pm0.29}$ & $\underline{45.66\pm0.69}$ & $\underline{55.13\pm0.80}$
& $13.94\pm0.27$ & $33.36\pm0.41$ & $\underline{45.01\pm0.40}$
& $\underline{27.94\pm0.60}$ & $\underline{55.15\pm0.90}$ & $\underline{66.25\pm0.53}$ \\
KME-no-log
& $\mathbf{24.94\pm0.12}$ & $\mathbf{47.14\pm0.16}$ & $\mathbf{56.69\pm0.34}$
& $\mathbf{17.14\pm0.32}$ & $\mathbf{37.66\pm0.35}$ & $\mathbf{48.96\pm0.42}$
& $\mathbf{32.05\pm0.66}$ & $\mathbf{58.37\pm0.30}$ & $\mathbf{68.62\pm0.39}$ \\
\bottomrule
\end{tabular}%
}
\end{table}

\begin{table}[htbp]
\centering
\caption{\blue{Matched zero-shot classification results after CC3M pretraining with
global batch size $512$. Values are percentages reported as mean $\pm$ sample
standard deviation over three seeds. Mean per-class accuracy is used for
Caltech-101, Aircraft, Flowers, and Pets; top-1 accuracy is used otherwise.
Average is computed over the 13 datasets within each seed before aggregation
across seeds. Bold indicates the highest mean and underlining indicates the
second-highest mean in each row.}}
\label{table:ablation-zero-shot}
\begin{tabular}{@{}lcccc@{}}
\toprule
Dataset & CLIP & WPSE & KME-CLIP & KME-no-log \\
\midrule
\textbf{Average}
& $26.89\pm0.86$ & $27.24\pm1.73$ & $\underline{28.68\pm0.96}$
& $\mathbf{29.18\pm0.17}$ \\
\midrule
ImageNet
& $19.61\pm0.27$ & $19.50\pm1.23$ & $\underline{21.16\pm0.32}$
& $\mathbf{22.72\pm0.32}$ \\
CIFAR-10
& $60.75\pm6.26$ & $60.20\pm7.53$ & $\underline{61.79\pm5.89}$
& $\mathbf{64.48\pm2.89}$ \\
CIFAR-100
& $27.88\pm0.67$ & $26.98\pm2.20$ & $\underline{29.89\pm2.28}$
& $\mathbf{32.66\pm1.26}$ \\
STL-10
& $84.60\pm0.49$ & $\underline{84.73\pm0.28}$ & $84.30\pm1.99$
& $\mathbf{84.92\pm1.34}$ \\
Food-101
& $13.15\pm0.99$ & $13.16\pm0.78$ & $\mathbf{15.15\pm1.05}$
& $\underline{14.72\pm1.55}$ \\
Caltech-101
& $50.07\pm1.40$ & $49.18\pm3.16$ & $\underline{50.92\pm0.61}$
& $\mathbf{54.91\pm0.97}$ \\
Cars
& $1.06\pm0.12$ & $0.95\pm0.09$ & $\mathbf{1.30\pm0.23}$
& $\underline{1.13\pm0.15}$ \\
Aircraft
& $0.97\pm0.10$ & $\mathbf{1.27\pm0.39}$ & $\underline{1.16\pm0.25}$
& $1.11\pm0.21$ \\
Flowers
& $11.80\pm0.68$ & $12.17\pm1.38$ & $\underline{12.88\pm0.84}$
& $\mathbf{13.26\pm0.60}$ \\
EuroSAT
& $15.89\pm5.22$ & $\underline{23.55\pm6.34}$ & $\mathbf{25.99\pm8.06}$
& $20.16\pm2.72$ \\
DTD
& $\underline{14.25\pm1.10}$ & $12.31\pm0.30$ & $13.79\pm1.24$
& $\mathbf{15.16\pm1.63}$ \\
Pets
& $11.65\pm0.36$ & $12.80\pm0.64$ & $\underline{14.31\pm0.22}$
& $\mathbf{15.07\pm0.90}$ \\
SUN397
& $37.84\pm0.98$ & $37.29\pm1.15$ & $\mathbf{40.25\pm0.58}$
& $\underline{39.00\pm0.87}$ \\
\bottomrule
\end{tabular}
\end{table}

\subsection{Known-PMI synthetic experiments}
\label{app:known-pmi}
To evaluate PMI fidelity directly, we use a continuous extension of the
2-mixture model in Section~\ref{section: theory_clip}. Each observation has an
unordered latent-class pair $\{i,j\}$ and $t\sim\mathrm{Unif}(0,1)$, represented
by $a_x=t e_i+(1-t)e_j$.
Here, $e_i\in\mathbb{R}^N$ denotes the $i$-th standard basis vector.
The latent vector $a_x$ defines the data distribution and the ground-truth
PMI but is not directly used as a fixed Transformer embedding; the Transformer
instead receives the two class indices and their scalar masses and learns its
own class-token embeddings.

We also consider independent-pair contamination: concretely, with probability $1-\eta$ we draw $(x,y)$ from the
shared-latent joint distribution, and with probability $\eta$ we draw
$x$ and $y$ independently from their marginals.
With contamination probability $\eta=0.1$, the
population density ratio is known analytically:
\[
    \exp(\operatorname{PMI}_\eta(x,y))
    =\eta+(1-\eta)N\langle a_x,a_y\rangle.
\]
We compare CLIP, WPSE, KME-no-log, and KME-CLIP for
$N\in\{8,16,32\}$, $d\in\{2,4,8,16\}$, and three seeds, using the same
one-layer shared Transformer (width $16$, two heads). We train on $32{,}768$
pairs for $16$ epochs with batch size $512$.
KME-no-log is matched to KME-CLIP
in parameterization---the same architecture and point-set size, the same
positive weights, the same Gaussian kernel, and the same initialization,
optimizer, seeds, and data---but uses $K_\theta(x,y)$ rather than
$\log K_\theta(x,y)$ as its similarity.
All learned Gaussian-kernel methods
use $\lambda=b^2$ with the common fixed initialization $b_0=1$.
All synthetic experiments in this section were run on a 2024
Mac mini with an Apple M4 Pro chip and 24~GB of unified memory.
We report a symmetric
conditional-KL estimate using fresh marginal candidate banks of size $1{,}024$
and a calibrated relative $L^2$ error of $\exp(\operatorname{PMI})$ on
independent samples. The conditional-KL quantity is, equivalently, the excess
of the contrastive risk over the analytic-PMI oracle; the figure axes label it
as such, and we write \emph{conditional KL} below.

We note one limitation of this design. With $\eta=0.1$ the density ratio
satisfies $\exp(\operatorname{PMI}_\eta)\ge\eta>0$, so $\operatorname{PMI}_\eta$
is bounded below and the regime discussed in
Remark~\ref{remark: wpse-theory} (where $p(x,y)=0$ forces
$\operatorname{PMI}\to-\infty$) does not arise. A similarity of the form
$K_\theta$ with a positive kernel and positive weights is itself strictly
positive and bounded below, whereas $\log K_\theta$ is not, so among the settings
covered by our assumptions this contaminated model is the one least favorable to
the outer logarithm. The comparison below should be read with that in mind.

\begin{figure}[htbp]
    \centering
    \includegraphics[width=\linewidth]
        {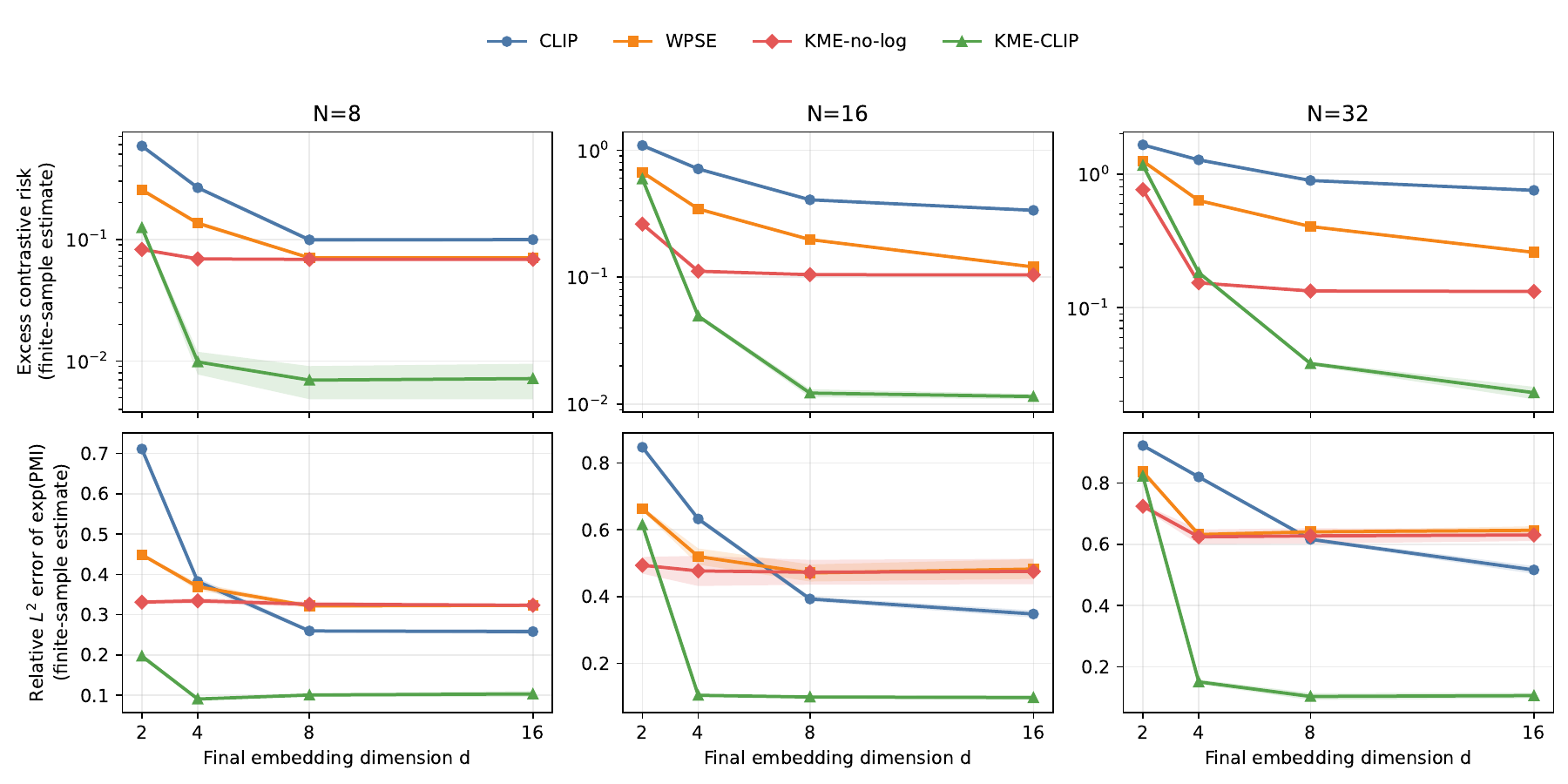}
    \caption{\blue{Known-PMI synthetic comparison across final embedding
    dimensions. Lines and shaded regions show mean and sample standard
    deviation over three seeds. Top: a finite-sample estimate of
    conditional KL, equivalently the excess contrastive risk over the
    analytic-PMI oracle, as labeled on the axis (lower is better; logarithmic
    axis). Bottom: a finite-sample
    estimate of the calibrated relative $L^2$ error of
    $\exp(\operatorname{PMI})$ (lower is better).}}
    \label{fig:known-pmi-synthetic}
\end{figure}

\paragraph{Fidelity at $\eta=0.1$.}
Figure~\ref{fig:known-pmi-synthetic} shows a dimension-dependent result.
KME-no-log has lower conditional KL at
$(N,d)\in\{(8,2),(16,2),(32,2),(32,4)\}$, while KME-CLIP is lower in the
other eight cells. KME-CLIP has lower relative $L^2$ error in all cells except
$(16,2)$ and $(32,2)$. Taken together, for every tested $N$ and for both
$d=8$ and $d=16$, KME-CLIP attains the lowest value among all four methods on
both fidelity metrics; the cells favoring KME-no-log are confined to $d\le4$.
Thus, the outer logarithm yields better values on both fidelity metrics at
$d\in\{8,16\}$, but not uniformly across all tested dimensions.
The next two paragraphs vary the contamination level to test whether this pattern is driven by the boundedness of $\operatorname{PMI}_\eta$, and evaluate the induced ranking against the analytic optimum.

\paragraph{Contamination sweep.}
The design above fixes $\eta=0.1$, the regime least favorable to the outer
logarithm: the density ratio is bounded below by $\eta$, so the divergence
discussed in Remark~\ref{remark: wpse-theory} does not occur. We therefore
repeat the experiment at $N=32$ over $\eta\in\{0.1,0.03,0.01,0.003\}$ and
$d\in\{2,4,8,16\}$, with the same three seeds $\{0,1,2\}$ and an otherwise
identical protocol; the $\eta=0.1$ column reuses the corresponding runs.
Because latent-class pairs are unordered pairs of distinct classes drawn
uniformly, a fraction $\binom{N-2}{2}/\binom{N}{2}=87.7\%$ of independently
drawn pairs sits exactly at the contamination floor
$\exp(\operatorname{PMI}_\eta)=\eta$. Decreasing $\eta$ therefore concentrates
the target near zero without ever reaching it: even at $\eta=0.003$ we have
$\operatorname{PMI}_\eta\ge\log\eta\approx-5.8$, so this is a trend study
towards the regime of Remark~\ref{remark: wpse-theory} rather than a
realization of it.

\begin{figure}[htbp]
\centering
\includegraphics[width=\linewidth]{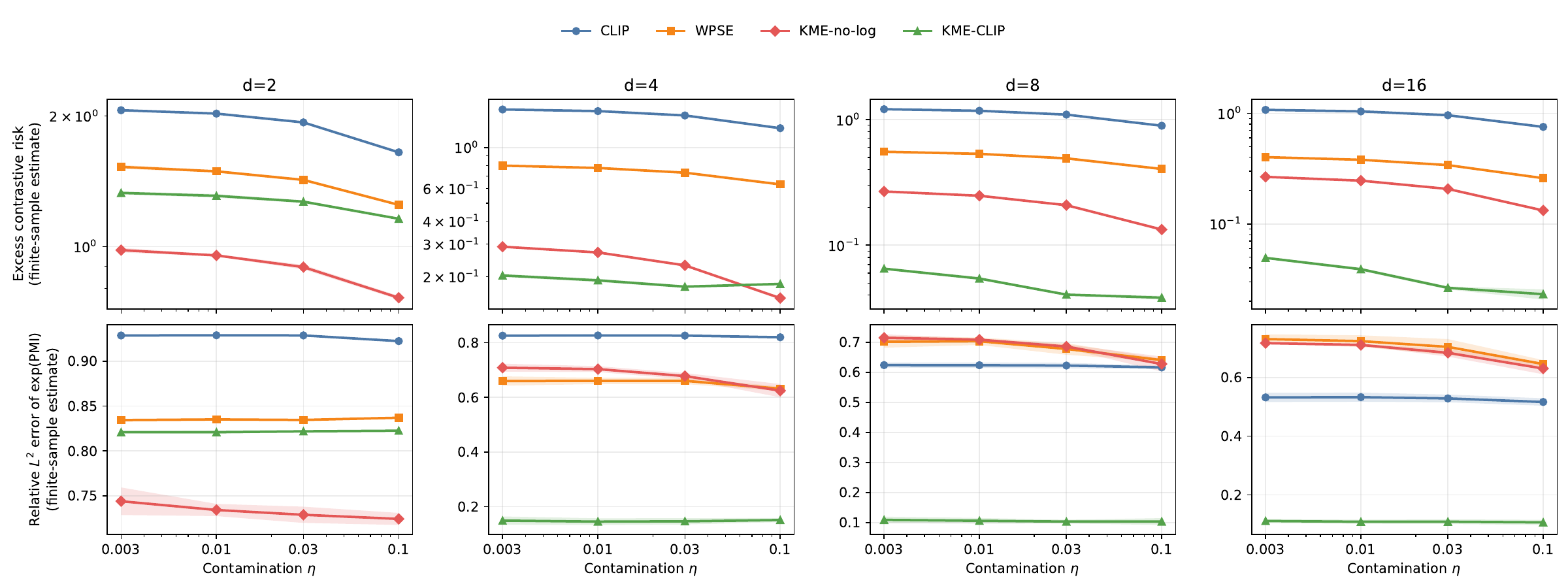}
\caption{\blue{Contamination sweep at $N=32$ across final embedding dimensions.
Lines and shaded regions show the mean and sample standard deviation over three
seeds. Top: conditional KL, equivalently the excess contrastive risk over the
analytic-PMI oracle, as labeled on the axis (lower is better; logarithmic
axis). Bottom: calibrated relative $L^2$ error of $\exp(\operatorname{PMI})$
(lower is better). The horizontal axis is $\eta$, decreasing to the left. The
conditional-KL ordering between KME-CLIP and KME-no-log reverses at $d=4$
between $\eta=0.1$ and $\eta=0.03$, whereas at $d=2$ KME-no-log remains better
throughout.}}
\label{fig:eta-sweep}
\end{figure}

\Cref{fig:eta-sweep} shows the two fidelity metrics and
\Cref{table:eta-gap} reports the within-seed paired differences. On the
relative $L^2$ error, KME-CLIP is essentially invariant to contamination at
$d\ge4$, its error changing by a factor of $0.99$--$1.05$ from $\eta=0.1$ to
$\eta=0.003$, whereas KME-no-log degrades by $1.13$--$1.14$; CLIP is flat only
because it already sits at an error level of $0.52$--$0.62$. On conditional KL
the informative case is $d=4$, where the ordering reverses in favor of
KME-CLIP as soon as $\eta\le0.03$. At $d=8$ and $d=16$ KME-CLIP attains the
lowest conditional KL at every $\eta$, but we do not read the widening
additive gap there as differential robustness: on the logarithmic scale the
ratio between the two methods is not monotone in $\eta$.
The exception is $d=2$, where KME-no-log is better on both metrics at every
$\eta$. \Cref{thm: kme-clip-superiority} asserts the existence of suitable
embeddings at every $d\ge2$ and says nothing about which of them training
recovers, so we report this as an empirical limitation at the smallest tested
embedding dimension without offering a mechanism.

\begin{table}[htbp]
\centering
\caption{\blue{Contamination sweep at $N=32$: within-seed paired difference
(KME-no-log minus KME-CLIP), reported as mean $\pm$ sample standard deviation
over three seeds. Positive values favor KME-CLIP. The paired difference is
computed seed by seed before aggregation.}}
\label{table:eta-gap}
\begin{tabular}{@{}lcccc@{}}
\toprule
& \multicolumn{4}{c}{Final embedding dimension $d$} \\
\cmidrule(lr){2-5}
$\eta$ & $2$ & $4$ & $8$ & $16$ \\
\midrule
\multicolumn{5}{@{}l}{Conditional KL} \\
$0.1$ & $-0.397 \pm 0.003$ & $-0.030 \pm 0.005$ & $+0.095 \pm 0.004$ & $+0.109 \pm 0.005$ \\
$0.03$ & $-0.373 \pm 0.008$ & $+0.054 \pm 0.006$ & $+0.168 \pm 0.004$ & $+0.181 \pm 0.005$ \\
$0.01$ & $-0.355 \pm 0.007$ & $+0.080 \pm 0.006$ & $+0.193 \pm 0.004$ & $+0.207 \pm 0.005$ \\
$0.003$ & $-0.348 \pm 0.003$ & $+0.088 \pm 0.007$ & $+0.203 \pm 0.005$ & $+0.217 \pm 0.005$ \\
\midrule
\multicolumn{5}{@{}l}{Relative $L^2$ error of $\exp(\operatorname{PMI})$} \\
$0.1$ & $-0.098 \pm 0.007$ & $+0.473 \pm 0.017$ & $+0.524 \pm 0.015$ & $+0.524 \pm 0.011$ \\
$0.03$ & $-0.093 \pm 0.010$ & $+0.531 \pm 0.017$ & $+0.582 \pm 0.010$ & $+0.575 \pm 0.018$ \\
$0.01$ & $-0.087 \pm 0.008$ & $+0.558 \pm 0.002$ & $+0.602 \pm 0.009$ & $+0.602 \pm 0.008$ \\
$0.003$ & $-0.077 \pm 0.017$ & $+0.559 \pm 0.013$ & $+0.606 \pm 0.016$ & $+0.606 \pm 0.004$ \\
\bottomrule
\end{tabular}
\end{table}

\paragraph{Ranking against the analytic Bayes ranker.}
Downstream retrieval measures the ordering a similarity induces, not its
calibration, and in \Cref{app:real-data-nolog} the two come apart. In the
present setting we can ask whether they also come apart when the best
achievable ordering is known, because the analytic density ratio yields an
optimal ranker.

The protocol is as follows. For a held-out pair $(x,y^+)$ we form a candidate
set consisting of $y^+$ together with $1{,}023$ items drawn i.i.d.\ from the
marginal, rank all $1{,}024$ candidates by the similarity under evaluation, and
record whether $y^+$ appears among the top $k$. We report the accuracy
averaged over the two retrieval directions, and use the same candidate-bank
size as in the conditional-KL estimate above.

We include the analytic density ratio $\exp(\operatorname{PMI}_\eta)$ as a
Bayes reference. It is optimal for this protocol. Writing $J$ for the index of
the positive in a candidate set $y_1,\dots,y_K$, exactly one of which is drawn
from $p(\cdot\mid x)$ while the rest are drawn from the marginal, we have
\[
    \Pr(J=j\mid x,y_1,\dots,y_K)
    \;\propto\;
    p(y_j\mid x)\prod_{i\neq j}p(y_i)
    \;\propto\;
    \frac{p(y_j\mid x)}{p(y_j)}
    \;=\;
    \exp(\operatorname{PMI}(x,y_j)),
\]
since $\prod_{i}p(y_i)$ does not depend on $j$. Ranking by the density ratio is
therefore ranking by the posterior probability of being the positive, and it
maximizes the probability that $y^+$ appears among the top $k$ for every $k$.
This is the same identity that makes PMI the minimizer of the contrastive loss
in \Cref{thm: pmi-minimzer}.

The reference is thus a ceiling rather than a competing method, and the ceiling
lies far below $100\%$: the $87.7\%$ of pairs that sit at the contamination
floor all share the density ratio $\eta$, and no ranker can separate them. The
task is nevertheless non-saturating and its ordering becomes better determined
as contamination decreases, the reference attaining a symmetric top-$10$
accuracy of $18.82\%$ at $\eta=0.1$ and $20.95\%$ at $\eta=0.003$
(\Cref{table:bayes-ranking}).

At $d\ge4$, KME-CLIP attains between $98.4\%$ and $99.8\%$ of the Bayes
reference top-10 accuracy at every contamination level, while KME-no-log
saturates between $89.3\%$ and $94.4\%$. The pattern is sharper at top-1: at
$d=16$ the reference improves from $2.03\%$ to $2.21\%$ as $\eta$ decreases
and KME-CLIP tracks it ($2.03\%$ to $2.23\%$; the paired ratio is $1.00$
within seed-to-seed variation, and values marginally above one reflect
finite-sample fluctuation rather than a claim of exceeding the optimum),
whereas KME-no-log remains at $1.46\%$--$1.48\%$ throughout. As the problem
becomes cleaner, the pre-log score fails to follow the improvement in the
attainable ranking, which is the behavior expected of a strictly positive,
bounded-below similarity when the target ratio concentrates near zero.
The
exception is again $d=2$. In this controlled setting, then, better
density-ratio fidelity and better ranking coincide, and the outer logarithm is
preferable on both; because the ground-truth density ratio is unavailable on
CC3M, we do not infer the cause of the reversal there.

\begin{table}[htbp]
\centering
\caption{\blue{Hard retrieval at $N=32$ with one held-out positive and
$1{,}023$ marginal negatives. Symmetric top-10 accuracy (\%), reported as mean $\pm$
sample standard deviation over three seeds. ``Bayes'' is the analytic
density-ratio ranker and gives the attainable ceiling under this protocol.}}
\label{table:bayes-ranking}
\begin{tabular}{@{}llccccc@{}}
\toprule
$d$ & $\eta$ & CLIP & WPSE & KME-no-log & KME-CLIP & Bayes \\
\midrule
\multirow{4}{*}{$4$}
& $0.1$ & $12.40 \pm 0.26$ & $16.16 \pm 0.34$ & $17.04 \pm 1.47$ & $18.73 \pm 0.33$ & $18.82 \pm 0.10$ \\
& $0.03$ & $13.25 \pm 0.40$ & $18.21 \pm 1.03$ & $18.91 \pm 1.24$ & $20.17 \pm 0.18$ & $20.44 \pm 0.49$ \\
& $0.01$ & $14.13 \pm 0.54$ & $18.83 \pm 0.82$ & $18.98 \pm 1.98$ & $20.72 \pm 0.12$ & $20.87 \pm 0.60$ \\
& $0.003$ & $14.21 \pm 0.22$ & $18.46 \pm 0.82$ & $18.72 \pm 1.03$ & $20.74 \pm 0.12$ & $20.95 \pm 0.56$ \\
\midrule
\multirow{4}{*}{$8$}
& $0.1$ & $16.73 \pm 0.25$ & $16.76 \pm 0.58$ & $17.69 \pm 0.34$ & $18.57 \pm 0.69$ & $18.82 \pm 0.10$ \\
& $0.03$ & $18.36 \pm 0.63$ & $17.95 \pm 0.33$ & $18.88 \pm 0.79$ & $20.39 \pm 0.38$ & $20.44 \pm 0.49$ \\
& $0.01$ & $18.83 \pm 1.18$ & $18.31 \pm 0.56$ & $19.60 \pm 0.81$ & $20.62 \pm 0.20$ & $20.87 \pm 0.60$ \\
& $0.003$ & $18.88 \pm 0.98$ & $18.44 \pm 0.66$ & $19.78 \pm 0.72$ & $20.65 \pm 0.17$ & $20.95 \pm 0.56$ \\
\midrule
\multirow{4}{*}{$16$}
& $0.1$ & $17.51 \pm 1.08$ & $16.89 \pm 0.73$ & $17.43 \pm 0.47$ & $18.52 \pm 0.03$ & $18.82 \pm 0.10$ \\
& $0.03$ & $18.65 \pm 1.26$ & $18.73 \pm 0.59$ & $18.86 \pm 1.26$ & $20.30 \pm 0.12$ & $20.44 \pm 0.49$ \\
& $0.01$ & $19.03 \pm 1.01$ & $19.22 \pm 1.30$ & $19.27 \pm 1.39$ & $20.61 \pm 0.20$ & $20.87 \pm 0.60$ \\
& $0.003$ & $19.17 \pm 0.74$ & $19.22 \pm 1.54$ & $19.42 \pm 1.38$ & $20.70 \pm 0.35$ & $20.95 \pm 0.56$ \\
\bottomrule
\end{tabular}
\end{table}

\end{document}